\documentclass[times,twocolumn,final]{elsarticle}

\usepackage{medima}
\usepackage{framed,multirow}

\usepackage{amssymb}
\usepackage{latexsym}

\usepackage{url}
\usepackage[dvipsnames]{xcolor}

\usepackage[colorlinks]{hyperref}

\usepackage{bbding}
\usepackage{soul}
\usepackage{bibunits}
\usepackage{bm}
\usepackage{natbib}
\usepackage{array}
\usepackage{subfigure}
\usepackage{pdflscape}
\usepackage{makecell}
\usepackage{booktabs}
\usepackage{framed,multirow}
\usepackage{threeparttable}
\usepackage{amssymb}
\usepackage{pifont}
\usepackage{algorithm}
\usepackage{layouts}
\usepackage{listings}
\usepackage{colortbl}
\usepackage{pythonhighlight}
\usepackage{amsmath}
\usepackage{enumitem}
\makeatletter
\newcommand\footnoteref[1]{\protected@xdef\@thefnmark{\ref{#1}}\@footnotemark}
\makeatother

\newcolumntype{P}[1]{>{\centering\arraybackslash}p{#1}}
\newlength\savewidth
\def\arrvline{\hfil\kern\arraycolsep\vline\kern-\arraycolsep\hfilneg}

\definecolor{Highlight}{HTML}{39b54a}  

\definecolor{iblue}{rgb}{0.03, 0.4, 1}
\definecolor{igray}{rgb}{0.00, 0.00, 0.00}

\definecolor{ForestGreen}{RGB}{34,139,34}

\newcommand{\oursupervisedmodel}{{\fontfamily{ppl}\selectfont SuPreM}}
\newcommand{\ourdataset}{{\fontfamily{ppl}\selectfont
AbdomenAtlas}}
\newcommand{\ourtrainpublic}{{\fontfamily{ppl}\selectfont
AbdomenAtlas 1.1}}
\newcommand{\ourtestprivateAA}{{\fontfamily{ppl}\selectfont
AbdomenAtlas-9K}}
\newcommand{\ourtestpublic}{{\fontfamily{ppl}\selectfont
FullBodyAtlas-1K}}
\newcommand{\ourchallenge}{{\fontfamily{ppl}\selectfont
BodyMaps}}
\newcommand{\numofct}{20,460}
\newcommand{\numofannotations}{673K}
\newcommand{\numofhospitals}{112}
\newcommand{\numofannotationslices}{7.6M}
\newcommand{\numofcountries}{19}

\definecolor{newcolor}{rgb}{.8,.349,.1}

\journal{Medical Image Analysis}

\begin{document}

\verso{Wenxuan Li \textit{et~al.}}

\begin{frontmatter}

\title{AbdomenAtlas: A Large-Scale, Detailed-Annotated, \& Multi-Center Dataset for\\Efficient Transfer Learning and Open Algorithmic Benchmarking}

\author[1]{Wenxuan \snm{Li}}
\author[1]{Chongyu \snm{Qu}}
\author[2]{Xiaoxi \snm{Chen}}
\author[1,3,4]{Pedro R. A. S. \snm{Bassi}}
\author[5]{Yijia \snm{Shi}}
\author[1,6]{Yuxiang \snm{Lai}}
\author[7]{Qian \snm{Yu}}
\author[8]{Huimin \snm{Xue}}
\author[1]{Yixiong \snm{Chen}}
\author[9]{Xiaorui \snm{Lin}}
\author[9]{Yutong \snm{Tang}}
\author[9]{Yining \snm{Cao}}
\author[9]{Haoqi \snm{Han}}
\author[10]{Zheyuan \snm{Zhang}}
\author[10]{Jiawei \snm{Liu}}
\author[1]{Tiezheng \snm{Zhang}}
\author[11]{Yujiu \snm{Ma}}
\author[12]{Jincheng \snm{Wang}}
\author[13,14,15]{Guang \snm{Zhang}}
\author[1]{Alan \snm{Yuille}}
\author[1]{Zongwei \snm{Zhou}\corref{cor1}}
\cortext[cor1]{Correspondence to: Zongwei Zhou (\href{mailto:zzhou82@jh.edu}{\texttt{zzhou82@jh.edu}})}

\address[1]{Department of Computer Science, Johns Hopkins University}
\address[2]{Department of Bioengineering, University of Illinois Urbana-Champaign}
\address[3]{Alma Mater Studiorum - University of Bologna}
\address[4]{Center for Biomolecular Nanotechnologies, Istituto Italiano di Tecnologia}
\address[5]{LKS Faculty of Medicine, The University of Hong Kong}
\address[6]{Department of Computer Science, Southeast University}
\address[7]{Department of Radiology, Southeast University Zhongda Hospital}
\address[8]{Department of Medical Oncology, The First Hospital of China Medical University}
\address[9]{The Second Clinical College, China Medical University}
\address[10]{Department of Mechanical Engineering and the Laboratory of Computational Sensing and Robotics, Johns Hopkins University}
\address[11]{Center of Reproductive Medicine, Department of Obstetrics and Gynecology, Shengjing Hospital of China Medical University}
\address[12]{Radiology Department, the First Affiliated Hospital, School of Medicine, Zhejiang University}
\address[13]{Department of Health Management, The First Affiliated Hospital of Shandong First Medical University \& Shandong Provincial Qianfoshan Hospital}
\address[14]{Shandong Engineering Research Center of Health Management}
\address[15]{Shandong Institute of Health Management}

\received{1 May 2013}
\finalform{10 May 2013}
\accepted{13 May 2013}
\availableonline{15 May 2013}
\communicated{S. Sarkar}

\begin{abstract}

We introduce the largest abdominal CT dataset (termed \ourdataset) of \numofct\ three-dimensional CT volumes sourced from \numofhospitals\ hospitals across diverse populations, geographies, and facilities. \ourdataset\ provides \numofannotations\ high-quality masks of anatomical structures in the abdominal region annotated by a team of 10 radiologists with the help of AI algorithms. We start by having expert radiologists manually annotate 22 anatomical structures in 5,246 CT volumes. Following this, a semi-automatic annotation procedure is performed for the remaining CT volumes, where radiologists revise the annotations predicted by AI, and in turn, AI improves its predictions by learning from revised annotations. Such a large-scale, detailed-annotated, and multi-center dataset is needed for two reasons. Firstly, \ourdataset\ provides important resources for AI development at scale, branded as \textit{large pre-trained models}, which can alleviate the annotation workload of expert radiologists to transfer to broader clinical applications. Secondly, \ourdataset\ establishes a large-scale benchmark for evaluating AI algorithms---the more data we use to test the algorithms, the better we can guarantee reliable performance in complex clinical scenarios. An ISBI \& MICCAI challenge named \textit{\ourchallenge: Towards 3D Atlas of Human Body} was launched using a subset of our \ourdataset, aiming to stimulate AI innovation and to benchmark segmentation accuracy, inference efficiency, and domain generalizability. We hope our \ourdataset\ can set the stage for larger-scale clinical trials and offer exceptional opportunities to practitioners in the medical imaging community. Codes, models, and datasets are available at \href{https://www.zongweiz.com/dataset}{\texttt{https://www.zongweiz.com/dataset}}

\end{abstract}


\end{frontmatter}

\section{Introduction}\label{sec:introduction}

\begin{table*}[t]
\centering
\scriptsize
    \caption{Our \ourdataset\ consists of three component datasets---\ourtrainpublic, \ourtestpublic, and \ourtestprivateAA, providing a total of \numofct\ annotated 3D computed tomography (CT) volumes, with many more to follow from a variety of sources. In \ourdataset\ (exclude JHH), we employed an efficient semi-automatic annotation procedure, described in \S\ref{sec:semi_automatic_annotation}, to annotate 25 anatomical structures for 15,214 CT volumes;
    in JHH, a team of expert radiologists provided very high-quality annotations for 22 anatomical structures for 5,246 CT volumes as illustrated in \figureautorefname~\ref{fig:abdomenatlas_property}--Property II.
    \ourtrainpublic\ will be made available to the public for AI training, \ourtestpublic\ is already publicly available for algorithmic benchmarking, and the CT volumes and annotations in \ourtestprivateAA\ will be reserved for rigorous external validation. Moreover, we invite further collaborations to expand detailed annotations on more CT volumes using our efficient human-AI synergy. It is generally agreed in the community that large-scale, detailed-annotated, and multi-center datasets are critical for AI benchmarking---the more data we use to test the algorithms, the better we can guarantee good performance in real-world conditions (e.g., clinical settings).}\vspace{2px}
    
\begin{tabular}{p{0.02\linewidth}p{0.2\linewidth}P{0.08\linewidth}P{0.09\linewidth}P{0.08\linewidth}P{0.06\linewidth}P{0.20\linewidth}P{0.085\linewidth}
}
\toprule
\multicolumn{2}{l}{\makecell[tl]{\ourdataset\ components}} & \makecell{\# of volumes\\(original)} & \makecell{\# of volumes\\(accessible)} & \makecell{\# of annotated \\structures} & \makecell{\# of\\hospitals} & \makecell{source countries} & \makecell{annotators} \\
\midrule
\multicolumn{8}{c}{\textit{purpose: AI training}} \\
\rowcolor{iblue!10}\multicolumn{2}{l}{\ourtrainpublic\ (public)} & 9,262
& 9,262 & 25 & 88 & \makecell{MT, IE, BR, BA, AUS, TH,\\ CA, TR, CL, ES, MA, US, \\DE, NL, FR, IL, CN } & human \& AI \\
 & CHAOS \citeyearpar{valindria2018multi} [\href{https://chaos.grand-challenge.org/Download/}{link}] & 40  & 20  & 1 & 1 & TR & human \\
 & BTCV \citeyearpar{landman2015miccai} [\href{https://www.synapse.org/#!Synapse:syn3193805/wiki/89480}{link}] & 50  & 47  & 12 & 1 & US & human \\
 & Pancreas-CT \citeyearpar{roth2015deeporgan} [\href{https://academictorrents.com/details/80ecfefcabede760cdbdf63e38986501f7becd49}{link}]            & 82  & 42  & 1 & 1 & US & human \\
 & CT-ORG \citeyearpar{rister2020ct} [\href{https://wiki.cancerimagingarchive.net/pages/viewpage.action?pageId=61080890#61080890cd4d3499fa294f489bf1ea261184fd24}{link}] & 140 & 140 & 6 & 8 & DE, NL, CA, FR, IL, US & human \& AI \\
 & WORD \citeyearpar{luo2021word} [\href{https://github.com/HiLab-git/WORD}{link}] & 150 & 120 & 16 & 1 & CN & human\\
 & LiTS \citeyearpar{bilic2019liver} [\href{https://competitions.codalab.org/competitions/17094}{link}]  & 201 & 130 & 2 & 7 & DE, NL, CA, FR, IL & human \\
 & AMOS22 \citeyearpar{ji2022amos} [\href{https://amos22.grand-challenge.org}{link}] & 500 & 200 & 15 & 2 & CN & human \& AI \\
 & KiTS \citeyearpar{heller2023kits21} [\href{https://kits-challenge.org/kits23/}{link}] & 600 & 489 & 3 & 1 & US & human \\
 & AbdomenCT-1K \citeyearpar{ma2021abdomenct} \citeyearpar{ma2023unleashing} [\href{https://github.com/JunMa11/AbdomenCT-1K}{link}] & 1,062 & 1,000 & 4 & 12 & DE, NL, CA, FR, IL, US, CN & human \& AI \\
 & MSD-CT \citeyearpar{antonelli2021medical} [\href{https://decathlon-10.grand-challenge.org/}{link}]           & 1,420 & 945 & 9 & 1 & US & human \& AI \\
 & FLARE’23 \citeyearpar{ma2022fast} [\href{https://codalab.lisn.upsaclay.fr/competitions/12239}{link}]   & 4,100 & 4,100 & 13 & 30 & - & human \& AI \\
 & Abdominal Trauma Det \citeyearpar{rsna-2023-abdominal-trauma-detection} [\href{https://www.rsna.org/education/ai-resources-and-training/ai-image-challenge/abdominal-trauma-detection-ai-challenge}{link}]   & 4,711 & 4,711 & 0 & 23 & \makecell{CL, DE, ES, TR, AUS, TH, \\MA, MT, CA, IE, BR, BA} & - \\
 
\midrule
\multicolumn{8}{c}{\textit{purpose: external validation}} \\
\rowcolor{iblue!10}\multicolumn{2}{l}{\ourtestpublic\ (public)} & 1,761 & 1,761 & 117 & 9 & CH, DE & human \& AI \\
 & DAP Atlas \citeyearpar{jaus2023towards} [\href{https://drive.google.com/file/d/1ex0a9eQULLvKPDwijmijX2h49A-ockNy/view}{link}] & 533 & 533 & 142 & 2 & DE & AI\\
 & TotalSegmentator \citeyearpar{wasserthal2022totalsegmentator} [\href{https://doi.org/10.5281/zenodo.6802613}{link}]       & 1,228 & 1,228 & 117 & 7 & CH & human \& AI \\
 
\rowcolor{iblue!10}\multicolumn{2}{l}{\ourtestprivateAA\ (private)} &  11,223 & 9,437 & 25 & 15 & US, CN, DE & human \& AI \\
 & Pancreas-CT (test) \citeyearpar{roth2015deeporgan} [\href{https://academictorrents.com/details/80ecfefcabede760cdbdf63e38986501f7becd49}{link}] & 82 & 38 & 1 & 1 & US & human \\
 & AMOS22 (test) \citeyearpar{ji2022amos} [\href{https://amos22.grand-challenge.org}{link}] & 500 & 300 & 0 & 2 & CN & human \& AI \\
 & AutoPET \citeyearpar{gatidis2022whole} [\href{https://wiki.cancerimagingarchive.net/pages/viewpage.action?pageId=93258287}{link}] & 1,014 & 445 & 0 & 2 & DE & - \\
 & MSD-CT (official test) \citeyearpar{antonelli2021medical} [\href{https://decathlon-10.grand-challenge.org/}{link}] & 1,420 & 465 & 0 & 1 & US & - \\
 & YF & 1,224  & 1,224 & 0   & 1 & CN & - \\
 & CirrhosisPro \citeyearpar{yu2023multimodal} & 1,737 & 1,719 & 2 & 7 & CN  & human \\
 & JHH \citeyearpar{xia2022felix} & 5,246 & 5,246 & 22 & 1 & US & human \\
\bottomrule
\end{tabular}
\begin{tablenotes}
    \item US: United States \quad DE: Germany \quad NL: Netherlands \quad CA: Canada \quad FR: France \quad IL: Israel \quad IE: Ireland \quad BR: Brazil \quad BA: Bosnia and Herzegowina
    \item CN: China \quad TR: Turkey \quad CH: Switzerland \quad AUS: Australia \quad TH: Thailand \quad CL: Chile \quad ES: Spain \quad MA: Morocco \quad MT: Malta 
\end{tablenotes}
\label{tab:abdomenatlas_makeup}
\end{table*}

Large pre-trained models have revolutionized natural language processing (NLP) with examples like GPTs \citep{brown2020language} and LLaMA \citep{touvron2023llama}. However, the road map to achieve such transformative models remains unfolding in computer vision (CV) despite fervent explorations being undertaken. The current strategies in CV are diverse: using pixels only (e.g., LVM \citep{bai2023sequential}), combining pixels with texts (e.g., LLaVA \citep{liu2023visual}), or incorporating detailed human annotations (e.g., SAM \citep{kirillov2023segment}). While these strategies have shown promise, they have yet to match the success level of language models that can be widely applicable across target tasks. This variety reflects the inherent complexity and varied requirements in processing image data compared with text data \citep{zhang2023challenges}.

A consensus is that \textit{large pre-trained models must be trained on massive, diverse datasets} \citep{moor2023foundation,blankemeier2024merlin}. The road we must take is to prepare massive, diverse datasets---and it would be even better if they were annotated. For language models, very large and diverse datasets are fairly easy to obtain (e.g., 250B web pages in the Common Crawl repository\footnote{Common crawl repository: \href{https://commoncrawl.org/}{\texttt{https://commoncrawl.org/}}}). For vision models, we are still very far from having a data source of comparable size and diversity (e.g., 5.85B images in the LAION-5B dataset \citep{schuhmann2022laion}). In particular, medical vision, bearing some resemblance to computer vision, is relatively new in this exploration and pretty much a vacuum in the search for sizable datasets \citep{chen2022recent}, especially for the most dominant 3D medical images \citep{blankemeier2024merlin}. Moreover, a unique issue of medical images extends to variations in data collections, imaging protocols, and patient demographics \citep{mckinney2020international,singh2022generalizability}, which is often overlooked in most existing datasets, as summarized in \tableautorefname~\ref{tab:abdomenatlas_makeup} and \S\ref{sec:review_public_dataset}. This raises a pressing concern about the generalizability of the pre-trained models.

This paper does not intend to discuss how to create GPT-like vision models in medical imaging but endeavors to provide the required data and annotations that could catalyze such discussions. We have collected and annotated \textbf{\numofct} CT volumes, totaling \textbf{\numofannotations} high-quality masks of anatomical structures in the abdominal region. These CT volumes are taken from \textbf{\numofhospitals} hospitals in \textbf{\numofcountries} countries, making this effort unprecedented in scale. It is, by far, the most extensive annotated medical dataset for AI benchmark and promises to be a valuable asset for the development of large pre-trained models in the medical domain. We name this dataset \ourdataset. A large dataset from diverse centers is needed for two main reasons: (I) The performance of AI algorithms is known to improve when they are trained on more data; the more data we use to test the algorithms, the better we can guarantee good performance under real-world conditions (e.g., clinical settings). (II) It is critically important to train and test AI algorithms on data from different centers because AI researchers have found that algorithms trained on data from one center may fail to generalize to data from other centers (as exemplified in  \citet{DeGrave2021AI} and \citet{Geirhos2020Shortcut}).

In the remainder of this paper, we begin with a review of the preexisting medical datasets that are publicly available and highlight the unique properties of our \ourdataset\ in \S\ref{sec:related work}. We then describe in depth the construction of \ourdataset\ in \S\ref{sec:abdomenatlas_construction}, elaborating time-consuming manual annotation for 5,246 CT volumes and efficient semi-automatic annotation procedure for the remaining 15,214 CT volumes. Following this, two practical applications of our \ourdataset\ are presented. Firstly, \S\ref{sec:application_transfer_learning} introduces a suite of large pre-trained models (\oursupervisedmodel) enabling efficient transfer learning across numerous downstream tasks, with a special analysis on transfer learning efficiency and ability. Secondly, \S\ref{sec:application_benchmark} describes an international competition (\ourchallenge) in collaboration with ISBI and MICCAI offering open algorithmic benchmarking AI reliability, efficiency, and generalizability in medical image segmentation. Finally, \S\ref{sec:discussion} concludes with a discussion of the current limitation and future promises of establishing large-scale, detailed-annotated, and multi-center datasets in medical image analysis.

\begin{figure*}[!h]
\centering
\includegraphics[width=1.0\linewidth]{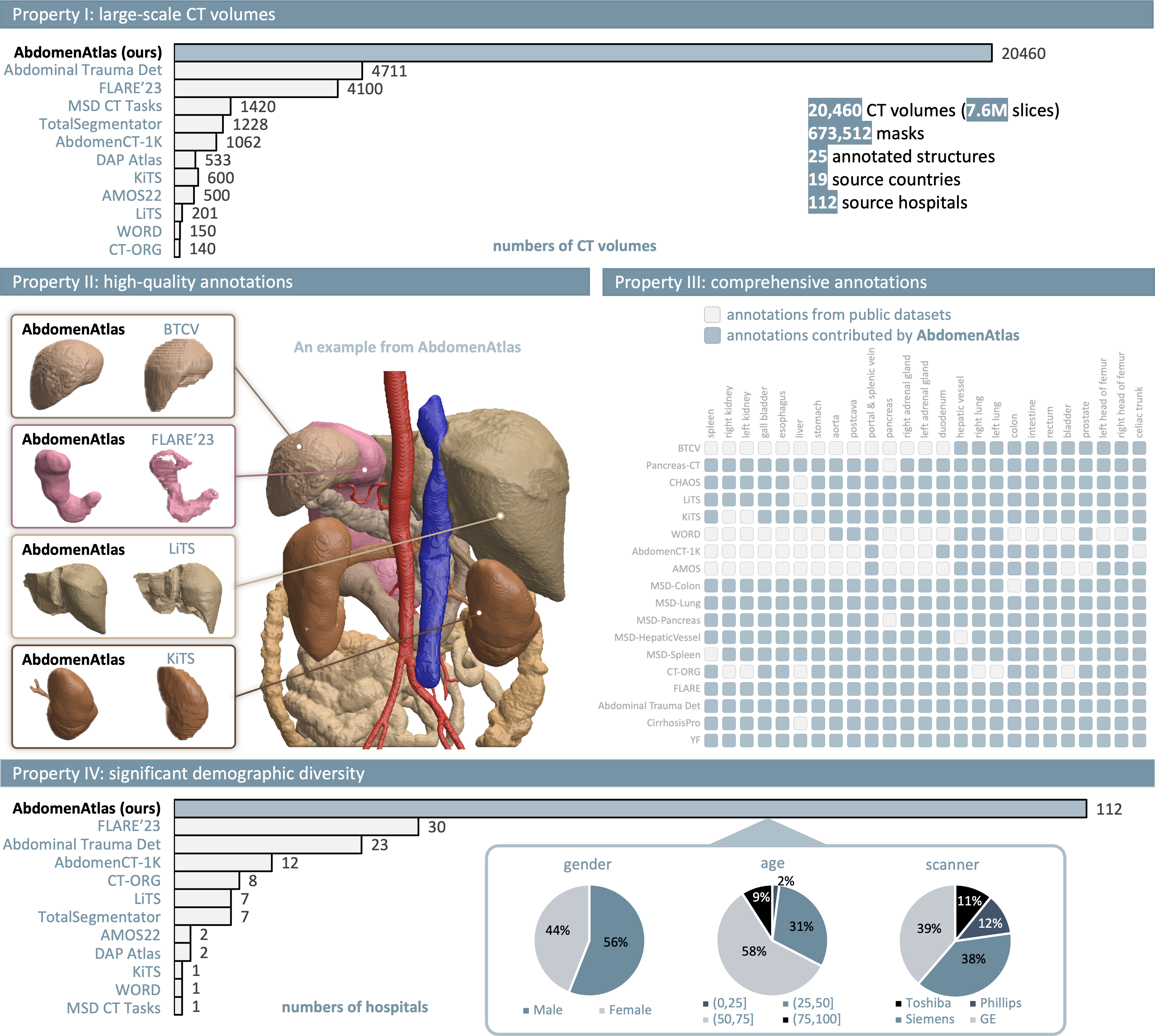}
    \caption{
    \textit{\ourdataset\ and related datasets.} Our \ourdataset\ stands out from other abdominal CT datasets in four unique properties: its unprecedented scale, superior quality of annotations, comprehensive nature of these annotations, and the broad demographic diversity it includes. Property I: \ourdataset\ provides the largest collection of annotated CT volumes among public radiology datasets, setting it apart in terms of scale. Property II: \ourdataset\ surpasses other public radiology datasets in the quality of its annotations, offering precise boundaries and accurate representation of each anatomical structure. Property III: \ourdataset\ provides detailed per-voxel annotations for 25 anatomical structures across each CT volume. Property IV: With CT volumes collected from \numofhospitals\ hospitals worldwide, \ourdataset\ also showcases an extensive diversity in the variety of CT scanners used and patient demographics, such as sex and age.
    }
\label{fig:abdomenatlas_property}
\end{figure*}

\section{\ourdataset\ and Related Datasets}\label{sec:related work}

\subsection{Preexisting Public Datasets}\label{sec:review_public_dataset}

\subsubsection{Classical Datasets ($<$500 CT Volumes)}

Classical datasets can be categorized into two groups. \textit{Group 1}: those designed for specific pathological conditions, such as LiTS (liver tumors), KiTS (kidney tumors), and Pancreas-CT (pancreatic tumors). These datasets usually provide more CT volumes (100s) but only annotate a specific type of anatomical structure and tumor. \textit{Group 2}: those designed for general purposes, such as BTCV (12 structures), and WORD (16 structures). These datasets annotate more types of structures, but due to the annotation cost, they are usually of small size (10s). These classical datasets, reviewed in \tableautorefname~\ref{tab:abdomenatlas_makeup}, have been invaluable public sources for training and validating state-of-the-art AI algorithms. As a significant advancement, our \ourdataset\ offers $50\times$ more CT volumes and $5\times$ more anatomical structures (classes) than these classical datasets. 

\subsubsection{Abdominal Multi-Organ Segmentation (AMOS)}\label{sec:review_amos}

The AMOS dataset  \citep{ji2022amos} includes 500 CT volumes and 100 MRI scans from patients with various abdominal conditions and different CT scanners. It provides detailed annotations of 15 anatomical structures and is valuable for cross-modality learning. However, the data is from only two hospitals in Asia, and important structures like the intestine and colon are not annotated. In contrast, \ourdataset\ is much more extensive, featuring CT volumes from $47\times$ more hospitals across $19\times$ more countries and includes annotations for 10 additional abdominal structures.

\subsubsection{AbdomenCT-1K} 
\label{sec:review_abdomenct1k}

The AbdomenCT-1K dataset \citep{ma2021abdomenct} provides 1,112 CT volumes from 12 hospitals, integrating data from five existing datasets and newly acquired CT volumes. It includes multi-phase, multi-vendor, and multi-disease cases. However, it only annotated four structures (liver, kidney, spleen, pancreas). On the contrary, \ourdataset\ contains annotations for $6.25\times$ more abdominal structures, providing more comprehensive 3D human body representations. 

\subsubsection{Medical Segmentation Decathlon (MSD) CT} 

The MSD-CT dataset \citep{antonelli2021medical} includes 1,420 CT volumes with nine anatomical structures annotated across six segmentation tasks---making it valuable for developing generalizable medical image segmentation algorithms. Unlike the partially annotated MSD-CT dataset, \ourdataset\ is fully annotated. We provide approximately 34K new masks, which are 35$\times$ more than those provided in the original MSD-CT dataset, as highlighted in the \figureautorefname~\ref{fig:abdomenatlas_property}--Property III.

\subsubsection{TotalSegmentator V2}
\label{sec:review_totalsegmentator}

The TotalSegmentator dataset \citep{wasserthal2022totalsegmentator} includes 1,228 CT volumes, focusing on whole-body segmentation of 117 anatomical structures. Derived from the University Hospital Basel, it features diverse cases, covering different ages, pathologies, scanners, body parts, and sequences.

However, there are three main issues with TotalSegmentator: \textit{(1) CT Volumes Quality}:  The CT volumes in TotalSegmentator are of lower quality due to resizing from the original 512$\times$512 resolution down to approximately 300$\times$300 to ease the data transfer. This resizing process inevitably results in the loss of fine-grained information. In contrast, \ourdataset\ provides CT volumes in their original resolution, totaling 1.8 TB, and we have made every effort to ease the data transfer, such as using Huggingface, Dropbox, Google Drive, and Baidu Wangpan.
\textit{(2) Annotation Quality}: While TotalSegmentator only uses semi-automatic labeling, we manually annotated 5,246 CT volumes in \ourdataset\ before using semi-automatic methods for the remaining, representing a significantly greater effort. Moreover, the annotations in TotalSegmentator were revised by two radiologists with three and six years of experience respectively.
In contrast, our team includes a more experienced group of ten radiologists with three to 15 years of experience. 
The comparison between TotalSegmentator and \ourdataset\ for hard-to-segment anatomical structures, such as the colon, is exemplified in \figureautorefname~\ref{fig:annotation_standard}. \textit{(3) Label Generation Procedure}: The labels in TotalSegmentator were largely produced by a single nnU-Net re-trained continually as shown in Figure 1b in \citet{wasserthal2022totalsegmentator}. Depending solely on nnU-Net could introduce a potential label bias favoring the nnU-Net architecture. This is evidenced by findings where nnFormer, UNETR, and Swin UNETR were all outperformed by nnU-Net and models building upon nnU-Net in TotalSegmentator \citep{huang2023stu}. To mitigate this bias, \ourdataset\ employed three different architectures (Swin UNETR, U-Net, and nnU-net) during the semi-automatic annotation procedure.

\subsubsection{Fast, Low-resource, and Accurate oRgan and Pan-cancer sEgmentation (FLARE)}\label{sec:review_flare}

The FLARE'23 dataset \citep{ma2022fast} contains 4,100 CT volumes from over 30 hospitals, with annotations for 13 abdominal structures and one tumor class. However, only 2,200 volumes are partially annotated, and 1,900 have no annotations. This incomplete annotation is due to the dataset's assembly from various existing datasets, each focusing on specific abdominal structures or tumors. Unlike partially annotated FLARE'23, \ourdataset\ fully annotated \numofct\ CT volumes with 25 anatomical structures. Moreover, \ourdataset\ offers higher annotation quality as shown in the stomach example in \figureautorefname~\ref{fig:abdomenatlas_property}--Property II. Additionally, most annotated structures in FLARE'23 are large structures that are relatively easier to detect/segment by humans and AI. \ourdataset\ provides annotations for hard-to-segment anatomical structures, such as the hepatic vessel, intestine, and colon.

\subsection{Four Properties in Our \ourdataset}\label{sec:abdomenatlas_property}

\subsubsection{Property I: Large-Scale CT Volumes} 

\ourdataset\ provides \numofct\ annotated CT volumes as shown in \figureautorefname~\ref{fig:abdomenatlas_property}--Property I, associated with over \numofannotationslices\ annotated CT slices. Besides providing details about \ourdataset, \tableautorefname~\ref{tab:abdomenatlas_makeup} presents its components and subdivisions for training (\ourtrainpublic) and testing (\ourtestpublic\ and \ourtestprivateAA). \ourdataset not only represents a substantial increase in the medical data available for AI training but also serves as an extensive resource for AI benchmarking. In \ourdataset, 9,262 annotated CT volumes in \ourtrainpublic\ will be made available to the public for the development of AI algorithms, and 1,761 annotated CT volumes in \ourtestpublic\ have already been publicly available for algorithmic benchmarking, thanks to TotalSegmentator \citep{wasserthal2022totalsegmentator} and DAP Atlas \citep{jaus2023towards}. Moreover, we have assembled and annotated 9,437 CT volumes from 15 hospitals, termed \ourtestprivateAA, which will be reserved for rigorous external validation. The scale of \ourdataset---\numofct\ CT volumes and \numofannotations\ masks---allows for both the development and evaluation of AI algorithms that can apply to a wide range of medical imaging tasks.

\subsubsection{Property II: High-Quality Annotations}

Creating \numofannotations\ high-quality masks for 25 anatomical structures requires extensive medical knowledge---at least three years of training in anatomical structures, and significant annotation costs---each structure taking about one hour for a radiologist to annotate \citep{park2020annotated}. As outlined in \S\ref{sec:annotation_standard}, we established a rigorous annotation standard based on human sectional anatomy \citep{dixon2017human} to guide the radiologists in accurately annotating or revising each structure, ensuring quality control. The efficacy of this standard in maintaining our annotation quality is illustrated in \figureautorefname~\ref{fig:abdomenatlas_property}--Property II, where the annotations in our \ourdataset\ show precise boundaries and accurate segmentation of anatomical structures, compared with those in BTCV, FLARE'23, LiTS, and KiTS.

\subsubsection{Property III: Comprehensive Annotations}
    
As depicted in \figureautorefname~\ref{fig:abdomenatlas_property}--Property III, we provide comprehensive per-voxel annotations for 25 anatomical structures, ensuring a fully-labeled dataset rather than a partially-labeled one from a naive combination of public datasets. Notably, different from the combination which only contains 39K masks, our \ourtrainpublic\ provides 231K annotated structures masks for these CT volumes, substantially increasing the available masks by 5.9$\times$. This increase not only enhances the dataset's utility but also enables the large-scale, supervised (pre-)training of AI algorithms in medical imaging analysis.

\subsubsection{Property IV: Significant Demographic Diversity} 

\ourdataset\ is a multi-center dataset of pre, portal, arterial, venous, and delayed phase CT volumes collected from \numofhospitals\ global hospitals across eight countries. As detailed in~\figureautorefname~\ref{fig:abdomenatlas_property}--Property IV, \ourdataset\ demonstrates demographic diversity, with a balanced sex distribution of 56\% female and 44\% male patients and a wide age range. Notably, 58\% patients aged 25 to 50 years, 31\% from 50 to 75 years, 9\% under 25 years, and 2\% over 75 years. Moreover, \ourdataset\ includes CT volumes from diverse scanners, such as Siemens, GE, Philips, and Toshiba, and incorporates CT volumes from both 16-/64-slice MDCT and Dual-source MDCT. These diversities in terms of phase, hospitals, countries, demography, scanners, and scan types enrich \ourdataset, ensuring that AI algorithms developed with \ourdataset\ can effectively handle variations in structure appearances influenced by different imaging protocols or patient positioning, such as rotations along the vertical axis between 30 and 60 degrees. Studies demonstrated that training data diversity is a key for AI distributional robustness \citep{fang2022data}. Accordingly, the great diversity of \ourdataset\ contributes to developing robust, fair, and generalizable AI algorithms capable of adapting to the diverse settings found in real-world clinical environments.

\begin{figure*}[t]
\includegraphics[width=1.0\linewidth]{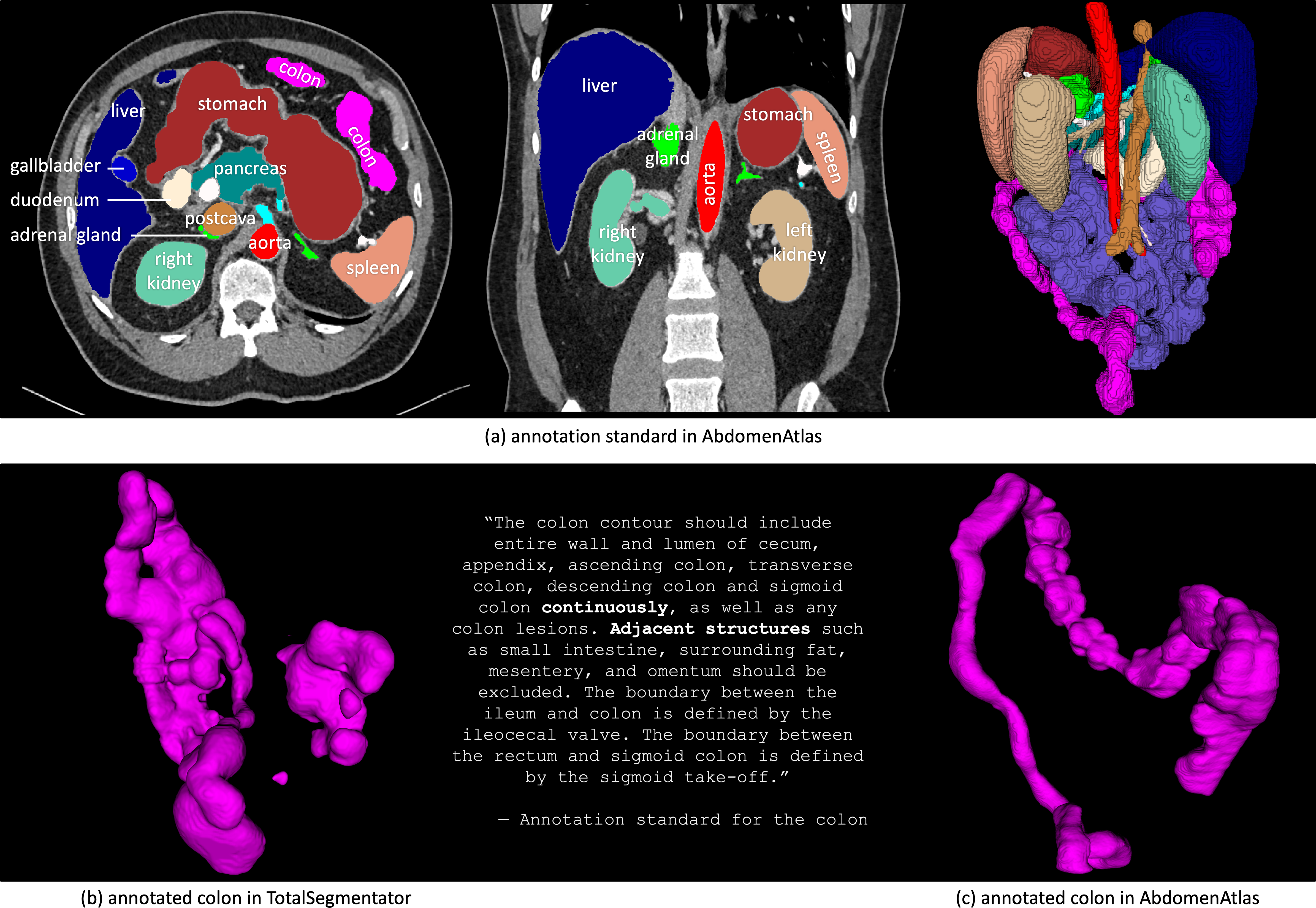}
    \caption{\textit{Our annotation standard and annotated colon (as an example) in public datasets.} (a) Our datasets are annotated according to strict standards either manually or semi-automatically. These annotation standards are strictly defined, to eliminate ambiguity in physiology and anatomy, ensuring uniform and standardized annotations. In contrast, many public datasets lack comprehensive and clear standards---especially for those hard-to-segment structures like the colon---resulting in ambiguous and erroneous annotations. This discrepancy is evident when comparing (b) and (c).
    }
\label{fig:annotation_standard}
\end{figure*}

\begin{table}[t]
\centering
\scriptsize
    \caption{\textit{Our study recruited ten radiologists, divided into two groups based on their experience.} The senior group consists of four radiologists with 8 to 15 years of experience in various specialties: one with 15 years in diagnostic radiology and gynecological diseases, another with 12 years in diagnostic radiology and abdominal/cerebral diseases, a third with 12 years in diagnostic radiology and thoracic diseases, and the last with 8 years in diagnostic radiology and abdominal diseases. The junior group includes six radiologists with 3 to 5 years of experience, all specializing in radiology. 
    }\vspace{2px}
    \centering
    \scriptsize
    \begin{tabular}{p{0.05\linewidth}P{0.10\linewidth}P{0.10\linewidth}P{0.55\linewidth}}
    \toprule
    & radiologist & experience & training/expertise\\ 
    \midrule
    \multirow{4}{*}{senior} & R1 & 15 years & diagnostic radiology, gynecological diseases\\
    & R2 & 12 years & diagnostic radiology, abdominal/cerebral diseases \\
    & R3 & 12 years & diagnostic radiology, thoracic diseases\\
    & R4 & 8 years & diagnostic radiology, abdominal diseases\\
    \midrule
    \multirow{6}{*}{junior} & R5 & 5 years & radiology\\	
    & R6 & 5 years & radiology\\
    & R7 & 4 years & radiology\\
    & R8 & 3 years & radiology\\
    & R9 & 3 years & radiology\\
    & R10 & 3 years & radiology\\
    \bottomrule
    \end{tabular}
\label{tab:annotator_info}
\end{table}

\section{Construction of \ourdataset}\label{sec:abdomenatlas_construction}

Annotation quality and consistency are our top priority in constructing \ourdataset. Therefore, we first established a comprehensive annotation protocol and standard (\S\ref{sec:annotation_standard}) designed to be reproducible by other teams when constructing similar datasets. Based on the standard, we applied two complementary annotation procedures. First, we adopted a manual annotation procedure---radiologists carefully annotated each CT volume voxel-by-voxel, ensuring high quality but requiring significant time investment (\S\ref{sec:manual_annotation}). Second, we adopted a semi-automatic annotation procedure combining radiologist expertise with AI algorithms (\S\ref{sec:semi_automatic_annotation})---radiologists revised AI predictions, guided by attention maps highlighting potential errors. This human-AI synergy increased the efficiency of creating large-scale, detailed annotated datasets by 168$\times$. Before release, four senior radiologists need to verify all the annotations in \ourdataset.

\subsection{Annotation Protocol and Standard}\label{sec:annotation_standard}

Our study recruited ten radiologists, including four senior radiologists with 8 to 15 years of experience and six junior radiologists with 3 to 5 years of experience. Detailed information is presented in \tableautorefname~\ref{tab:annotator_info}. For annotation accuracy and consistency, all radiologists familiarized themselves with the annotation standard as described below. We employed tools included a licensed version from \href{https://aipair.com.cn/}{Pair} and an open-source \href{https://www.slicer.org/}{3D Slicer} for annotation and revision. 
We provide the annotation standard for 25 structures in \ourdataset, including 16 abdominal organs (esophagus, stomach, duodenum, intestine, colon, rectum, liver, gall bladder, spleen, pancreas, left kidney, right kidney, left adrenal gland, right adrenal gland, bladder, prostate), 2 thorax organs (left lung, right lung), 5 vascular structures (aorta, celiac trunk, postcava, portal \& splenic vein, hepatic vessel), and 2 skeletal structures (left and right femur). An example of a detailed annotated CT volume is in \figureautorefname~\ref{fig:annotation_standard}.

\subsubsection{Abdominal Organs (Gastrointestinal Tract)} 

The \ul{stomach} contour should encompass the entire stomach wall and lumen including the fundus, body, antrum, and pylorus, as well as any gastric lesions, while adjacent structures, organs, and surrounding fat should be excluded. The \ul{duodenum} contour should include the entire duodenal wall and lumen from the duodenal bulb to the ligament of Treitz, along with any duodenal lesions, it should exclude surrounding structures such as the head of the pancreas, common bile duct, and surrounding vessels. The \ul{intestine} contour should include the jejunum and ileum wall and lumen from the ligament of Treitz to the ileocecal valve, along with any intestinal lesions, it should exclude surrounding fat, mesentery, and mesenteric vessels. The \ul{colon} contour should include the entire wall and lumen of the cecum, appendix, ascending colon, transverse colon, descending colon, and sigmoid colon, as well as any colon lesions, while adjacent structures, surrounding fat, mesentery, and omentum should be excluded. The \ul{rectum} contour should include the entire rectal wall, lumen, and any lesions, while adjacent structures, surrounding fat, and muscle should be excluded.

\subsubsection{Abdominal Organs (Others)} 

The \ul{liver} contour should include all the liver parenchyma and any lesions, the intrahepatic vessels and intrahepatic bile ducts need to be covered, while excluding surrounding fat, adjacent structures, and organs. The \ul{gallbladder} contour should encompass the entire gallbladder wall and lumen, including the fundus, body, and neck, as well as any gallstones or polyps, while the cystic duct, the surrounding liver parenchyma, and fat should be excluded. The \ul{pancreas} contour should encompass all pancreatic parenchyma including the head, body, and tail, as well as any pancreatic lesions and pancreatic duct, the surrounding vessels and fat should be excluded. The \ul{spleen} contour should include all splenic parenchyma and any lesions, it should exclude adjacent structures and extrasplenic vessels. The \ul{adrenal gland (L/R)} contour should include the entire adrenal gland and any adrenal lesions, it should exclude adjacent structures and surrounding fat. The \ul{kidney (L/R)} contour should include the renal parenchyma, excluding the renal pelvis, ureter, extrarenal blood vessels, surrounding fat, and any adjacent structures. The \ul{bladder} contour should include the entire adrenal wall, lumen, and any bladder lesions, it should exclude adjacent structures and surrounding fat. The \ul{prostate} contour should include the whole prostate parenchyma, prostatic urethra, and any prostate lesions, while excluding adjacent structures, surrounding fat, and prostatic venous plexus.

\subsubsection{Thorax Organ} 

The \ul{esophagus} contour should include the entire esophageal wall and lumen along with any esophageal lesions, while adjacent structures such as the trachea, aorta, and surrounding fat and muscle should be excluded. The \ul{lung (L/R)} contour should include the entire lung parenchyma, pulmonary broncho-vascular bundle, visceral pleura, and any pulmonary lesions. It should exclude pleural effusion, pneumothorax, parietal pleura, mediastinal structures, and chest wall.

\subsubsection{Vascular Structures}  

The \ul{aorta} and \ul{celiac trunk} contour should include the entire lumen of the arteries. The artery wall and calcification, ulcers, thrombosis, and dissection should also be included. The \ul{postcava} and \ul{portal \& splenic vein} contour should include the entire lumen and cover the walls, as well as intraluminal thrombus and tumor thrombus. The \ul{hepatic vessel} contour should include all intrahepatic vessel walls and lumen, as well as intraluminal thrombus and tumor thrombus.

\subsubsection{Skeletal Structures}

The \ul{femur (L/R)} contour should include the cortical bone and spongy bone, as well as any lesions. It should exclude surrounding muscles and vessels.

\subsection{Time-consuming Manual Annotation Procedure}\label{sec:manual_annotation}

 The manual annotation procedure involves radiologists annotating each CT volume voxel by voxel according to the annotation standard defined in \S\ref{sec:annotation_standard}. While this approach can ensure accuracy and consistency, reflecting the specific needs and requirements of the data, it is time-consuming, labor-intensive, and susceptible to human errors or biases. Annotation time for a single structure may range from minutes to hours, depending on the size and complexity of the regions of interest to annotate and the local surrounding anatomical structures \citep{park2020annotated}. This procedure was applied to annotate the JHH dataset: A total of 22 structures for each CT volume were annotated by a team of radiologists, and confirmed by one of three additional experienced radiologists, none of whom performed the annotations, to ensure the quality of the annotation \citep{xia2022felix}. JHH, involving 5,246 CT volumes, took years to complete and required the efforts of 15 radiologists.

 The precision of manual annotation in \ourdataset\ ensures that each anatomical structure is clearly defined, accurately capturing the hard-to-segment details of the body's physiology. As shown in \figureautorefname~\ref{fig:annotation_standard}, \ourdataset\ stands out in the precise manual annotation of hard-to-segment structures like the colon, demonstrating a clear advantage over datasets such as TotalSegmentator, which may have issues with annotations that erroneously include adjacent structures or are discontinuous. Despite these issues, TotalSegmentator remains a valued dataset as providing precise annotations for hard-to-segment structures is a rarity in public datasets. This is largely because manually annotating such structures is a meticulous and time-intensive task. None of the public datasets offer manual annotations for these structures across 5,246 CT volumes, setting \ourdataset\ apart regarding both scale and annotations.

  However, applying this approach to create annotations for the remaining 15,214 CT volumes in our \ourdataset\ is extremely time-consuming. Assuming an 8-hour workday over a five-day week, a trained radiologist generally requires 60 minutes to annotate each anatomical structure within a single CT volume \citep{park2020annotated}. Consequently, to annotate all 15,214 CT volumes, a radiologist would need 60$\times$25$\times$15,214 (minutes) /60/8/5 = 9,508 (weeks) = 182.9 (years). This motivated us to develop a more efficient annotation procedure.

\begin{figure*}[!t]
\includegraphics[width=\linewidth]{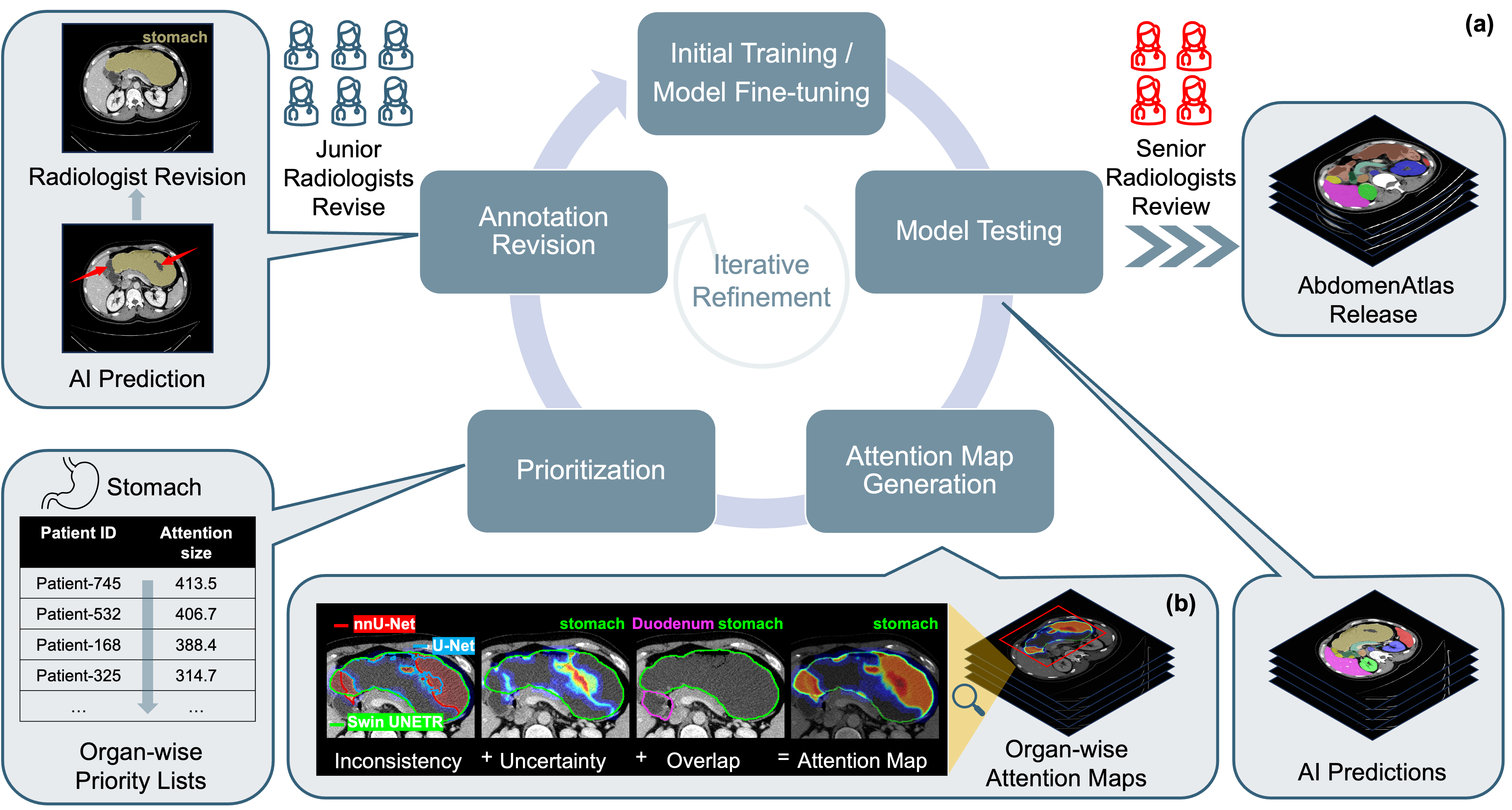}
    \caption{
    \textit{The semi-automatic annotation procedure} efficiently combines the best of radiologists with AI algorithms. In this procedure, AI predictions with potential errors are manually revised by six junior radiologists under the supervision of four senior radiologists. These precise revisions play a crucial role in fine-tuning the AI algorithms, leading to a continuous improvement in their performance. Consequently, the annotations created by the AI become increasingly reliable and robust, demonstrating a successful integration of human intelligence and artificial intelligence. \textit{(a) Semi-automatic annotation procedure overview.}
    In the seven-step cyclic semi-automatic annotation procedure, AI-predicted errors are highlighted using attention maps for each organ type. Subsequently, organ-wise priority lists are created, directing radiologists to focus on revising the most significant prediction errors, as signified by the largest attention map sizes. This prioritized approach ensures a more efficient and effective enhancement of the AI's predictive accuracy. \textit{(b) Attention map generation.} The attention map integrates three criteria: inconsistency (divergence in multiple AI predictions), uncertainty (high entropy value signaling low confidence in AI predictions), and overlap (intersection of multiple organ predictions). Therefore, the attention map highlights areas with the greatest potential for prediction errors.
    }
\label{fig:semi_automatic_annotation}
\end{figure*}

\subsection{Semi-automatic Annotation Procedure}\label{sec:semi_automatic_annotation}

The semi-automatic annotation procedure combines three different AI algorithms---to minimize any bias stemming from the model architectures---on public datasets of labeled CT volumes. These AI algorithms generate initial annotations for unlabeled CT volumes. We develop an innovative strategy that can find the most important sections of the AI predictions and use color-coordinated \textit{attention maps} to show radiologists which areas to focus on in their manual review of the AIs’ work \citep{qu2023annotating,li2024well}. As illustrated in \figureautorefname~\ref{fig:semi_automatic_annotation}, repeating this process---AI predictions and human review---over and over allowed us to accelerate the annotation process by an impressive factor of 168.

\subsubsection{Attention Maps Reveals AI Mistakes}\label{sec:attention_maps}

We develop attention maps to highlight potential errors in AI predictions, guiding radiologists during the review and revision process. These maps assign higher values to regions where the AI is more likely to have made mistakes, indicating areas for prioritized review. The attention map is computed using the following three criteria.

\begin{enumerate}
    \item \textbf{Inconsistency} is the standard deviation of soft predictions from three AI architectures, including Swin UNETR, nnU-Net, and U-Net. Regions with high standard deviation indicate high divergence in model predictions, prompting the need for additional manual revision.
    \begin{equation}
        \footnotesize
        \text{Inconsistency}_{i,c} = \sqrt{\frac{\sum_{n=1}^N(p_{i,c}^n-\mu_{i,c})^2}{N}},
    \end{equation}
    The subscript $c$ denotes class $c$ of annotated organs. For each voxel $i$, $p_{i,c}^n$ (ranging from 0 to 1) is the soft prediction value from the $n$-th AI architecture of class $c$ at that voxel's index $i$. $\mu_{i,c}$ represents the average prediction value obtained by combining results from three AI architectures at the same voxel index. With three AI architectures (denoted by $N=3$), $\text{Inconsistency}_{i,c}$ is determined by the standard deviation of the soft prediction values.

    \item \textbf{Uncertainty}, derived from AI soft predictions' entropy values, indicates areas with reduced confidence and heightened ambiguity, potentially increasing the risk of prediction errors in those regions.
    \begin{equation}
        \footnotesize
        \text{Uncertainty}_{i,c} = -\frac{\sum_{n=1}^N p_{i,c}^n\times\log(p_{i,c}^n)}{N}.
    \end{equation}
    The $\text{Uncertainty}_{i,c}$ is averaged over different AI architectures ($N=3$).

    \item \textbf{Overlap} signifies a prediction mistake based on organ prior, specifically when a voxel is predicted to belong to multiple organs, signaling an error even in the absence of ground truth information. 
    \begin{equation}
        \footnotesize
        \text{Overlap}_{i,c} = 
            \begin{cases}
                1 & \text{if}~p_{i,c}^n > 0.5 ~~\text{and}~\exists~ p^n_{i,c_{\notin}} > 0.5\\
                0 & \text{otherwise}
            \end{cases}
    \end{equation}
    $\text{Overlap}_{i,c}$ is determined as follows: if the prediction value for class $c$ exceeds 0.5 (a threshold value) for at least one AI architecture and there exists a prediction value not belonging to class $c$ that exceeds 0.5 for the same voxel index $i$, then the overlap value is set to 1; otherwise, it is set to 0.
\end{enumerate}

An attention map is created by combining inconsistency, uncertainty, and overlapping regions, facilitating annotators in efficiently identifying areas requiring revision or confirmation.
\begin{equation}
\footnotesize
\label{eq:attention}
    \text{Attention}_{i,c} = \text{Inconsistency}_{i,c} + \text{Uncertainty}_{i,c} + \text{Overlap}_{i,c}
\end{equation}
A higher $\text{Attention}_{i,c}$ value in the attention map indicates an increased risk of prediction errors for that voxel.

\subsubsection{Seven-Step Annotation Procedure}
\label{sec:step_by_step_procedure}

We employ a cyclical seven-step annotation procedure as illustrated in \figureautorefname~\ref{fig:semi_automatic_annotation}(a), which integrates human expertise with AI predictions. This procedure is designed to enhance AI performance gradually, leading to the generation of increasingly accurate and reliable annotations.

\begin{enumerate}
    \item \textbf{Initial Training.} An AI algorithm, denoted as $\mathcal{M}_0$ is trained from scratch utilizing 5,195 CT volumes sourced from 15 partially labeled public datasets.

    \item \textbf{Model Testing.} The models\footnote{Three models (i.e., Swin UNETR, U-Net, and nnU-Net) were used in this study to reduce architectural bias.}, $\mathcal{M}_0$, undergo direct inference on 15,214 CT volumes (include \ourtrainpublic, \ourtestpublic\ and \ourtestprivateAA, excluding those from JHH) to segment 25 anatomical structures.

    \item \textbf{Attention Map Generation.} For each CT volume, organ-wise attention maps are generated. These maps use a combination of inconsistency, uncertainty, and overlap metrics to identify regions potentially containing prediction errors.

    \item \textbf{Prioritizing Annotation Revision.} For each anatomical structure, a priority list is generated, where AI predictions are ranked across 15,214 CT volumes based on their \textit{attention size}, i.e. the cumulative intensity of \textit{attention map} across all voxels. A higher \textit{attention size} indicates an urgent requirement for manual revision due to a greater probability of prediction errors.

    \item \textbf{Performing Annotation Revision.} Six junior radiologists revise the AI predictions for the top 5\%\footnote{We empirically chose the threshold of 5\%. First, we analyzed the initial distribution of attention size for the CT volumes in \ourdataset. For most volumes, attention size was small. However, there were several significant-sized outliers, and the top 5\% of CT volumes with the highest attention size captured a large proportion of these outliers. Second, manually revising 5\% of the samples is feasible considering our annotation budget for each Step in the semi-automatic annotation procedure. If numerous outliers emerge or budgetary limitations exist, the threshold for revision priority should be re-calibrated.} of samples pertaining to each anatomical structure, as identified by the organ-specific priority lists, with the guidance of the organ-specific attention maps. The remaining 95\% of samples are retained without modification in Step (5). However, they will undergo automatic reassessment by the improved AI in the next cycle of the active learning process (Step 2). Samples with segmentation errors that were not selected for manual revision in an iteration of the semi-automatic annotation procedure should be selected in a future iteration, as the AI improves and other annotation errors are corrected.

    \item \textbf{Model Fine-Tuning.} Using only the manually revised annotations, the current AI algorithm, $\mathcal{M}_t$, is fine-tuned to yield an improved version, $\mathcal{M}_{t+1}$.

    \item \textbf{Iterative Refinement.} Steps \textit{(2)} through \textit{(6)} are repeated until the model's predictions for the most critical CT volumes are validated by annotators to require no further revisions, indicating minimal prediction errors.

\end{enumerate}

\noindent Prior to the dataset release, four senior radiologists are responsible for a comprehensive review of the entire \ourdataset~and make revisions\footnote{Such revisions are seldom required based on our study, with only about 100 out of 15,214 samples per anatomical structures need further adjustments.} if needed, particularly to address any potential errors that may remain after the semi-automatic annotation procedure is complete.

\begin{table*}[t]
    \centering
    \scriptsize
    \caption{\textit{A suite of pre-trained models (\oursupervisedmodel) includes several widely recognized AI models.} We offer pre-trained AI models, such as CNN, Transformer, and their combined versions, with plans to add more in the future. Each model was supervised pre-trained using large datasets and voxel-by-voxel annotations from \ourtrainpublic. Compared with learning from scratch and public models, fine-tuning the models in \oursupervisedmodel\ consistently leads to the state-of-the-art performance in organ/cardiac/vertebrae/muscle/tumor segmentation on two datasets, measured by Dice Similarity Coefficient (DSC) scores. Results provided with the mean and standard deviation (mean$\pm$s.d.) are across ten trials. Additionally, we further conducted an independent two-sample $t$-test comparing the results of learning from scratch and fine-tuning models in \oursupervisedmodel. The improvement in performance is statistically significant at the $P=0.05$ level, indicated by a \textcolor{iblue!75}{light blue} box. Here, \citet{tang2022self}, \citet{xie2022unimiss} and \citet{zhou2019models} represent self-supervised pre-trained models, and the remaining pre-trained models employed supervised pre-training.
    }
    \vspace{2px}
    \begin{tabular}{p{0.15\linewidth}p{0.06\linewidth}p{0.2\linewidth}|P{0.05\linewidth}P{0.05\linewidth}P{0.05\linewidth}P{0.05\linewidth}|P{0.05\linewidth}P{0.05\linewidth}P{0.05\linewidth}}
    \toprule
    \multicolumn{3}{l|}{} & \multicolumn{4}{c|}{TotalSegmentator} & \multicolumn{3}{c}{our proprietary dataset} \\
    pre-trained model & params & pre-trained data & organ & muscle & cardiac & vertebrae & organ & gastro & cardiac \\
    \midrule
     \multicolumn{3}{l|}{\textit{backbone: U-Net \citep{ronneberger2015u} and its variants}} & & & & & & &  \\
    scratch & 19.08M & none & 88.9{\tiny$\pm$0.6} & 92.9{\tiny$\pm$0.4} & 88.8{\tiny$\pm$0.7} & 86.9{\tiny$\pm$0.3} & 85.6{\tiny$\pm$0.5
    } & 69.8{\tiny$\pm$1.2} & 38.1{\tiny$\pm$1.1}\\
    \citet{zhou2019models} & 19.08M & 623 CT volumes & 87.8 & 90.1 & 86.3 & 85.1 & 80.1 & 65.5 & 36.9 \\
    \citet{chen2019med3d} & 85.75M & 1,638 CT volumes and masks & 86.9 & 91.4 & 87.4 & 82.2 & 79.0 & 66.2 & 36.7 \\
    \citet{xie2022unimiss} & 61.79M & 5,022 CT\&MRI volumes & 88.5 & 92.9 & 89.0 & 85.2 & - & - & -\\
    \citet{zhang2021dodnet} & 17.29M & 920 CT volumes and masks& 89.3 & 93.8 & 89.1 & 86.0 & 85.7 & 72.7 & 38.3\\
    \rowcolor{iblue!10}\oursupervisedmodel & 19.08M & 2,100 CT volumes and masks & 92.1{\tiny$\pm$0.3} & 95.4{\tiny$\pm$0.1} & 92.2{\tiny$\pm$0.3} & 91.3{\tiny$\pm$0.2} & 90.8{\tiny$\pm$0.2} & 76.2{\tiny$\pm$0.8} & 70.5{\tiny$\pm$0.5}\\ 
    \midrule
    \multicolumn{3}{l|}{\textit{backbone: Swin UNETR \citep{hatamizadeh2021swin} and its variants}} & & & & & & &  \\
    scratch & 62.19M & none & 86.4{\tiny$\pm$0.5} & 88.8{\tiny$\pm$0.5}& 84.5{\tiny$\pm$0.6} & 81.1 {\tiny$\pm$0.5}& 77.3{\tiny$\pm$0.9}& 65.9{\tiny$\pm$ 1.7}& 35.5{\tiny$\pm$1.4} \\
    \citet{tang2022self} & 62.19M & 5,050 CT volumes & 89.3 & 93.8 & 88.3 & 86.2 & 87.9 & 72.5 & 38.9\\
    \citet{liu2023clip,liu2024universal} & 62.19M & 2,100 CT volumes and masks & 89.7 & 94.1 & 89.4 & 86.5 & 89.1 & 74.6 & 67.6\\
    \rowcolor{iblue!10}\oursupervisedmodel & 62.19M & 2,100 CT volumes and masks & 91.3{\tiny$\pm$0.3} & 94.6{\tiny$\pm$0.2}& 90.3{\tiny$\pm$0.3}& 87.2{\tiny$\pm$0.3} & 90.4{\tiny$\pm$0.7} & 75.9{\tiny$\pm$1.2}& 69.8{\tiny$\pm$0.9}\\ 
    \midrule
    \multicolumn{3}{l|}{\textit{backbone: SegResNet \citep{myronenko20193d}}} & & & & & & & \\
    scratch & 4.7M & none & 88.6{\tiny$\pm$0.5}& 91.3{\tiny$\pm$0.4} & 89.8{\tiny$\pm$0.4} & 87.6{\tiny$\pm$0.2} & 80.6{\tiny$\pm$0.8} & 67.0{\tiny$\pm$1.4} & 36.0{\tiny$\pm$1.3}\\
    \rowcolor{iblue!10}\oursupervisedmodel & 4.7M & 2,100 CT volumes and masks & 91.3{\tiny$\pm$0.5}& 94.0{\tiny$\pm$0.1}& 91.3{\tiny$\pm$0.5} & 89.5{\tiny$\pm$0.2} & 86.6{\tiny$\pm$0.3} & 73.7{\tiny$\pm$1.0} & 67.9{\tiny$\pm$0.8}\\ 
    \bottomrule
    \end{tabular}
    \label{tab:model_zoo}
\end{table*}

\section{Application I: Efficient Transfer Learning}\label{sec:application_transfer_learning}

We present \oursupervisedmodel, a suite of pre-trained 3D models that learns generalizable representations from \ourdataset\ and provides a basis for annotation-efficient model adaptation in several applications. Specifically, \oursupervisedmodel\ is trained for semantic segmentation on
annotated CT volumes (from \ourtrainpublic) by means of supervised learning and then adapted to novel class segmentation and disease detection tasks with explicit annotations. Therefore, \oursupervisedmodel\ have the potential to serve as Foundation Models \citep{bommasani2021opportunities,moor2023foundation}. In this section, the transfer learning ability is assessed by segmentation performance on sub-datasets from \ourtestpublic\ and \ourtestprivateAA.

\subsection{A Suite of Pre-trained Models: \oursupervisedmodel}
\label{sec:suprem}

Our \ourdataset\ dataset stands out in terms of its extensive scale and comprehensive annotations, offering a significant benefit for training AI models through both supervised and self-supervised manner. As of the writing of this paper, there have been no instances of either supervised or self-supervised pre-training conducted on a dataset of this magnitude (3+ million images from 9,262 volumetric data)\footnote{For supervised pre-training, the largest research thus far was conducted by \citet{liu2023clip}, using a total of 3,410 annotated CT volumes, split into 2,100 for training and 1,310 for validation. On the other hand, the largest study in self-supervised pre-training was carried out by \citet{tang2022self}, employing 5,050 CT volumes that were unannotated. \textbf{Concurrently}, \citet{valanarasu2023disruptive} pre-trained a model on an even larger dataset, consisting of 50,000 volumes that included both CT and MRI images, using self-supervised learning.}. Leveraging the comprehensive scope of our \ourtrainpublic, we have developed a suite of models (termed \oursupervisedmodel). These models are built upon CNN backbones, such as U-Net and SegResNet, as well as Transformer backbones, such as Swin UNETR. As the use of pre-trained models becomes more widespread, there is an increasing need for standardized and easily accessible approaches for sharing public model weights. Accordingly, we have released a suite of pre-trained models summarized in~\tableautorefname~\ref{tab:model_zoo}. Releasing pre-trained foundation models should be considered a significant contribution, offering an alternative approach for knowledge sharing while simultaneously safeguarding patient privacy \citep{zhang2023challenges}.

To perform a fair and rigorous comparison between \oursupervisedmodel\ and state-of-the-art supervised and self-supervised pre-trained models, we limited the \oursupervisedmodel\ pre-training dataset to only 2,100 CT volumes from \ourtrainpublic. This size is the same as that in \citet{liu2023clip} and fewer than \citet{tang2022self} (\tableautorefname~\ref{tab:model_zoo}). \tableautorefname~\ref{tab:preliminary_challenge} shows results for training \oursupervisedmodel\ on the entire \ourtrainpublic\ and directly testing on \ourtestprivateAA. Benchmarking results indicate that, compared with learning from scratch and existing public models, models fine-tuned from \oursupervisedmodel\ consistently achieve better segmentation performance across both TotalSegmentator and our proprietary dataset. In this section, all of the models in \oursupervisedmodel\ follow pre-training and fine-tuning configurations as below.

\begin{itemize}
    
    \item \textbf{Pre-training:} 
    We use a random cropping method to extract sub-volumes of dimensions 96$\times$96$\times$96 voxels from the original CT volumes. Our \oursupervisedmodel\ is pre-trained on \ourtrainpublic\ configured with $\beta_1=0.9$, $\beta_2=0.999$, and a batch size of 2 per GPU, using AdamW optimizer and a cosine learning rate schedule with a warm-up for the first 100 epochs. \oursupervisedmodel\ starts with an initial learning rate of $1e^{-4}$ and a decay of $1e^{-5}$. The pre-training has been carried out on four NVIDIA A100 using multi-GPU (4) setup with distributed data-parallel (DDP), implemented in MONAI 0.9.0, with a maximum of 800 epochs. As for the objective function of pre-training, we use the binary cross-entropy and Dice Similarity Coefficient (DSC) losses. The selection of the best model is based on achieving the highest average DSC score, calculated across 25 classes on the validation set. 

    \item \textbf{Fine-tuning:} We fine-tune the pre-trained models using two sub-datasets from \ourdataset\ (i.e., the TotalSegmentator from \ourtestpublic\ and our proprietary dataset). During fine-tuning, we maintain the initial configurations from pre-training, while modifying the warm-up schedule to 20 epochs, setting a maximum of 200 epochs, and using a single GPU. For the fine-tuning, we employ cross-entropy and DSC loss as the objective function.
    
\end{itemize}

For all the other pre-training models compared in~\tableautorefname~\ref{tab:model_zoo}, we follow the recommended network architectures and hyper-parameter settings from their published papers for optimal performance. All the task performances are evaluated by the segmentation metric know as DSC.

\subsection{Efficiency in Transfer Learning}
\begin{figure*}[t]
\centering
\includegraphics[width=1.0\linewidth]{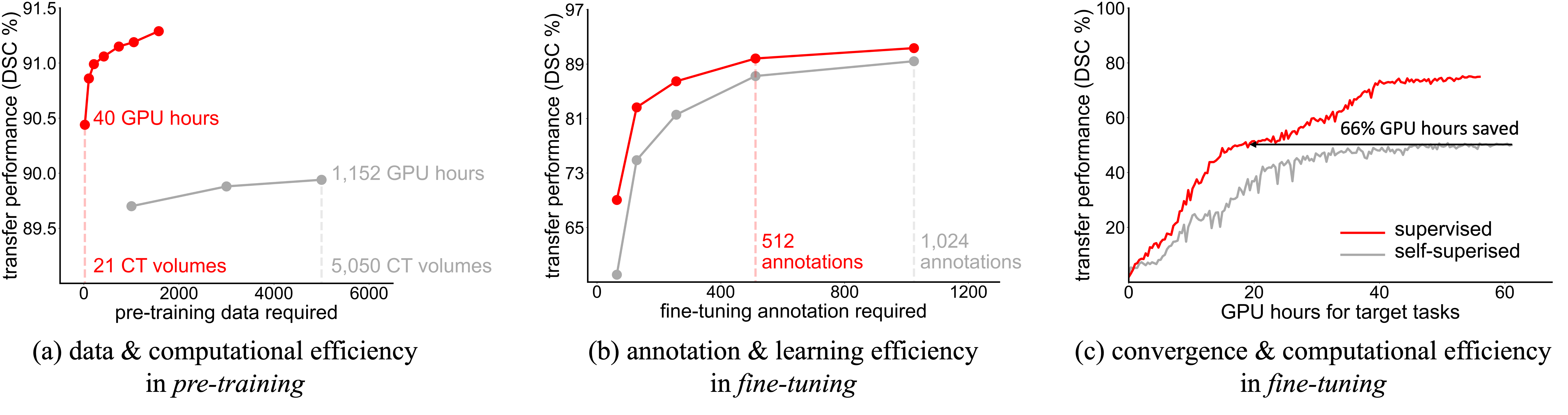}
    \caption{\textit{Data, annotation, convergence, and computational efficiency.} For a rigorous comparison, both supervised (shown in \textcolor{red}{red}) and self-supervised (shown in \textcolor{gray}{gray}) models use Swin UNETR as their backbone. The supervised pre-training is based on our \oursupervisedmodel\ while the self-supervised pre-training uses the current state-of-the-art model \citep{tang2022self}. \textit{(a)} The model's transfer learning ability improves consistently with an increase in the number of pre-training CT volumes, as evidenced by the results from TotalSegmentator. The model pre-trained with 21 CT volumes, 672 masks, and 40 GPU hours demonstrates a comparable transfer learning ability to that trained with 5,050 CT volumes and 1,152 GPU hours. Notably, supervised pre-training proves to be more efficient, requiring substantially 99.6\% fewer data and 96.5\% less computation in pre-training. \textit{(b)} \oursupervisedmodel\ requires 50\% fewer manual annotations in fine-tuning for novel organ/cardiac/vertebrae/muscle segmentation than self-supervised pre-training models. \textit{(c)} \oursupervisedmodel\ reaches convergence faster than the self-supervised pre-training model, leading to a 66\% GPU hours reduction in fine-tuning.
    }
\label{fig:suprem_efficiency}
\end{figure*}

In~\figureautorefname~\ref{fig:suprem_efficiency}, we present the notable efficiency of \oursupervisedmodel\ in transferring from pretext tasks to target tasks. Specifically, (1)  \oursupervisedmodel\ pre-trained with 21 CT volumes, 672 masks, and 40 GPU hours achieves a transfer learning ability comparable to that of a self-supervised model \citep{tang2022self} pre-trained with 5,050 CT volumes and 1,152 GPU hours. (2) \oursupervisedmodel\ requires 50\% fewer manual annotations in fine-tuning when transferring to target tasks of various anatomical structures segmentation than self-supervised pre-training. (3) \oursupervisedmodel\ converges much faster than the self-supervised pre-trained model, resulting in 66\% GPU hours saved when transferring to the target task. These results indicate the superior transfer learning efficiency of supervised pre-training with semantic segmentation.

\subsubsection{Data Efficiency}
As depicted in~\figureautorefname~\ref{fig:suprem_efficiency}(a), the need for data in \oursupervisedmodel\ is considerably lower (21 compared to 5,050 CT volumes) than in self-supervised pre-training. This difference stems from the inherent distinct learning objectives and the information used by them. Supervised pre-training (\oursupervisedmodel) gains advantages from explicit annotations, which directly guide the task, such as segmentation in this case. The model acquires knowledge from both the data and its annotations, receiving strong and precise supervision. In contrast, self-supervised learning depends on pretext tasks extracting learning features from the raw, unannotated data itself, which often results in a more unclear learning signal and necessitates a greater number of examples to capture valuable features. Notably, our results indicate that supervised pre-training scales better with increased data. When the data quantity is increased from 21 to 1,575 volumes, there is an improvement in transfer performance on TotalSegmentator, from 90.4\% to 91.3\%. In comparison, in self-supervised pre-training, expanding the dataset from 1,000 to 5,050 volumes only leads to a marginal improvement in performance, from 89.7\% to 89.9\%. Therefore, our \oursupervisedmodel\ not only requires substantially less data than self-supervised but also shows greater scalability and effectiveness when introducing more data.

\subsubsection{Annotation Efficiency}

We evaluated the annotation efficiency by fine-tuning \oursupervisedmodel\ and self-supervised models \citep{tang2022self}, using varying numbers of annotated CT volumes from TotalSegmentator. As indicated in \figureautorefname~\ref{fig:suprem_efficiency}(b), fine-tuning \oursupervisedmodel\ can lead to a 50\% reduction in manual annotation costs for anatomical structures segmentation, averaged across organs, muscles, cardiac and vertebrae which were not part of the pre-training classes. In particular, when \oursupervisedmodel\ is fine-tuned using 512 annotated CT volumes, it shows a similar transfer learning ability to \citet{tang2022self} fine-tuned with 1,024 annotated volumes. This improvement in fine-tuning performance becomes more pronounced as the number of annotated CT volumes available for the target task is limited (e.g., 64, 128, 256).

\subsubsection{Computational Efficiency}
\begin{figure}[h]
    \includegraphics[width=0.9\linewidth]{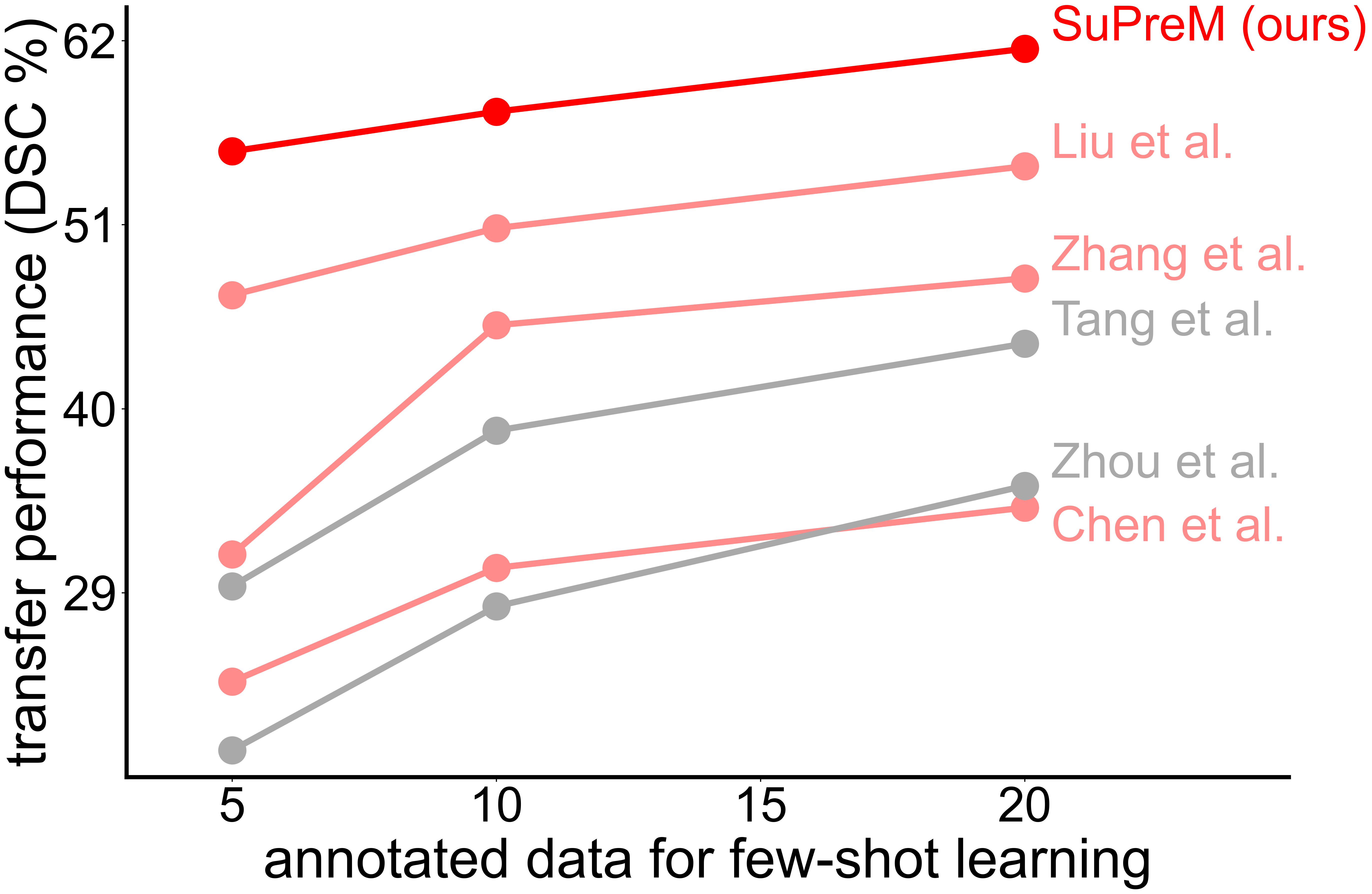}
    \caption{
    We exhibit the transfer learning ability of our \oursupervisedmodel\ on the proprietary dataset using a few examples ($N$ = $5,10,20$). Here, the transfer learning ability (represented on the Y-axis) is defined by the mean DSC score across the segmentation of 20 organ classes and 3 tumor classes. Generally, in scenarios of few-shot learning, models pre-trained with supervision (indicated in \textcolor{red}{red}) exhibit superior transfer ability compared to those pre-trained without supervision (shown in \textcolor{gray}{gray}). Importantly, our \oursupervisedmodel\ outperforms other widely recognized public models in terms of transfer learning ability. 
    }
    \label{fig:few_shot}
\end{figure}
The computational efficiency primarily arises from the reduced data demands inherent to supervised pre-training, as previously mentioned. As depicted in \figureautorefname~\ref{fig:suprem_efficiency}(a), \oursupervisedmodel\ requires only 40 GPU hours to match the transfer learning performance of self-supervised pre-training, which needs 1,152 GPU hours, equating to an increase by a factor of 28.8$\times$. In \figureautorefname~\ref{fig:suprem_efficiency}(c), \oursupervisedmodel\ reaches convergence more rapidly than the self-supervised pre-training for target tasks like fine-tuning on a 10\% subset of TotalSegmentator, decreasing the needed GPU hours from 60 to 20. This suggests that the image features learned through supervised pre-training are inherently more representative, allowing the model to effortlessly transfer to various segmentation tasks using minimal annotated data for fine-tuning. Such computational efficiency makes supervised pre-training an attractive option for target segmentation tasks, with robust model performance, particularly when an annotated dataset is available for pre-training, such as \ourdataset.

\subsection{Transfer Learning Ability}
\label{sec:transfer_learning_ability}
The transfer learning ability of \oursupervisedmodel\ shows significant generalization and adaptability through its learned features. These features can be fine-tuned to address few-shot scenarios, segment fine-grained pancreatic tumor classes, and classify tumor sub-types with higher accuracy compared to those learned by self-supervision.

\subsubsection{Transfer to Few-shot Scenarios}
\label{sec:few_shot_learning}

The exploration of few-shot learning scenarios is essential to assess the adaptability and efficiency of AI models in medical image analysis, particularly when annotated data is scarce. We present the transfer performance of publicly available models and \oursupervisedmodel\ on the proprietary dataset with a limited number (N) of annotated CT volumes (N = 5, 10, and 20). \figureautorefname~\ref{fig:few_shot} illustrates that \oursupervisedmodel\ consistently outperforms other state-of-the-art models across all few-shot scenarios. The transfer performance, measured by the DSC score, shows a significant improvement as the number of annotated CT volumes increases, with \oursupervisedmodel\ achieving the highest DSC at each few-shot scenario. This trend highlights the superior transfer learning abilities and robustness of \oursupervisedmodel\ to deal with data scarcity.

\begin{table}[t]
\centering
\scriptsize
    \caption{Fine-tuning \oursupervisedmodel\ on segmenting three novel fine-grained pancreatic tumor types from the subset of \ourtestprivateAA. Dice Similarity Coefficient (DSC) scores show that our \oursupervisedmodel, supervised pre-trained on 2,100 annotated data, demonstrates superior transfer learning ability to three novel fine-grained pancreatic tumor classes than the self-supervised model \citep{tang2022self} pre-trained on 5,500 raw, unlabeled data. In addition, we have further performed an independent two-sample $t$-test between \citet{tang2022self} and \oursupervisedmodel. The performance gain ($\Delta$) is statistically significant at the $P=0.05$ level, with highlighting in a \textcolor{iblue!75}{light blue} box. 
    }\vspace{2px}
    \centering
    \scriptsize
    \begin{tabular}{p{0.25\linewidth}P{0.25\linewidth}P{0.2\linewidth}P{0.1\linewidth}}
    \toprule
    novel class & \citet{tang2022self} & \oursupervisedmodel & $\Delta$ \\ 
    \midrule
    PDAC & 53.4{\tiny$\pm$0.3} & 53.6{\tiny$\pm$0.4} & 0.2\\		
    Cyst & 41.6{\tiny$\pm$0.4} & \cellcolor{iblue!10}49.2{\tiny$\pm$0.5} & 7.6\\		
    PanNet & 35.4{\tiny$\pm$0.8} & \cellcolor{iblue!10}45.7{\tiny$\pm$0.8} & 10.2\\
    \midrule
    \textbf{average (tumors)} & 48.9{\tiny$\pm$0.4} & \cellcolor{iblue!10}53.1{\tiny$\pm$0.4}& 4.2\\	
    \bottomrule
    \end{tabular}
\label{tab:novel_classes}
\end{table}

\subsubsection{Transfer to Novel Class Segmentation}
\label{sec:novel_class_segmentation}

The essence of transfer learning involves fine-tuning the pre-trained models to novel scenarios \citep{zhou2021models}, such as novel classes that are completely unseen during pre-training. In this study, we evaluate the transfer learning ability of \oursupervisedmodel\ when transferred to segment novel classes that are even challenging for expert radiologists. These novel classes are three fine-grained pancreatic tumor types from a
subset of \ourtestprivateAA\. As indicated in \tableautorefname~\ref{tab:novel_classes}, when transferred to these tumor classes, our \oursupervisedmodel, supervised pre-trained on anatomical structures, has better transfer learning ability than those self-supervised models pre-trained on raw, unlabeled data. We observe that the pretext task of segmentation itself inherently improves the model's ability to segment objects for novel classes. This advantage is more direct and comprehensible than the benefits offered by self-supervised tasks like contextual prediction, image masking, or instance discrimination, especially in transfer learning scenarios. We hypothesize that this is due to the supervised model learning to understand \textit{objectness}---a concept that defines an entity as an object within an image, setting it apart from the background or other entities. Through full supervision in segmentation tasks, the model develops a deeper understanding of what characterizes an object, extending beyond the recognition of specific objects to grasp fundamental object characteristics. These characteristics include texture, boundaries, shape, size, and other vital low-level visual elements crucial for basic image segmentation. 

\subsubsection{Transfer to Fine-grained Tumor Identification}
\label{sec:fine_grained_tumor_identification}

\begin{figure*}[t]
\centering
\includegraphics[width=0.9\linewidth]{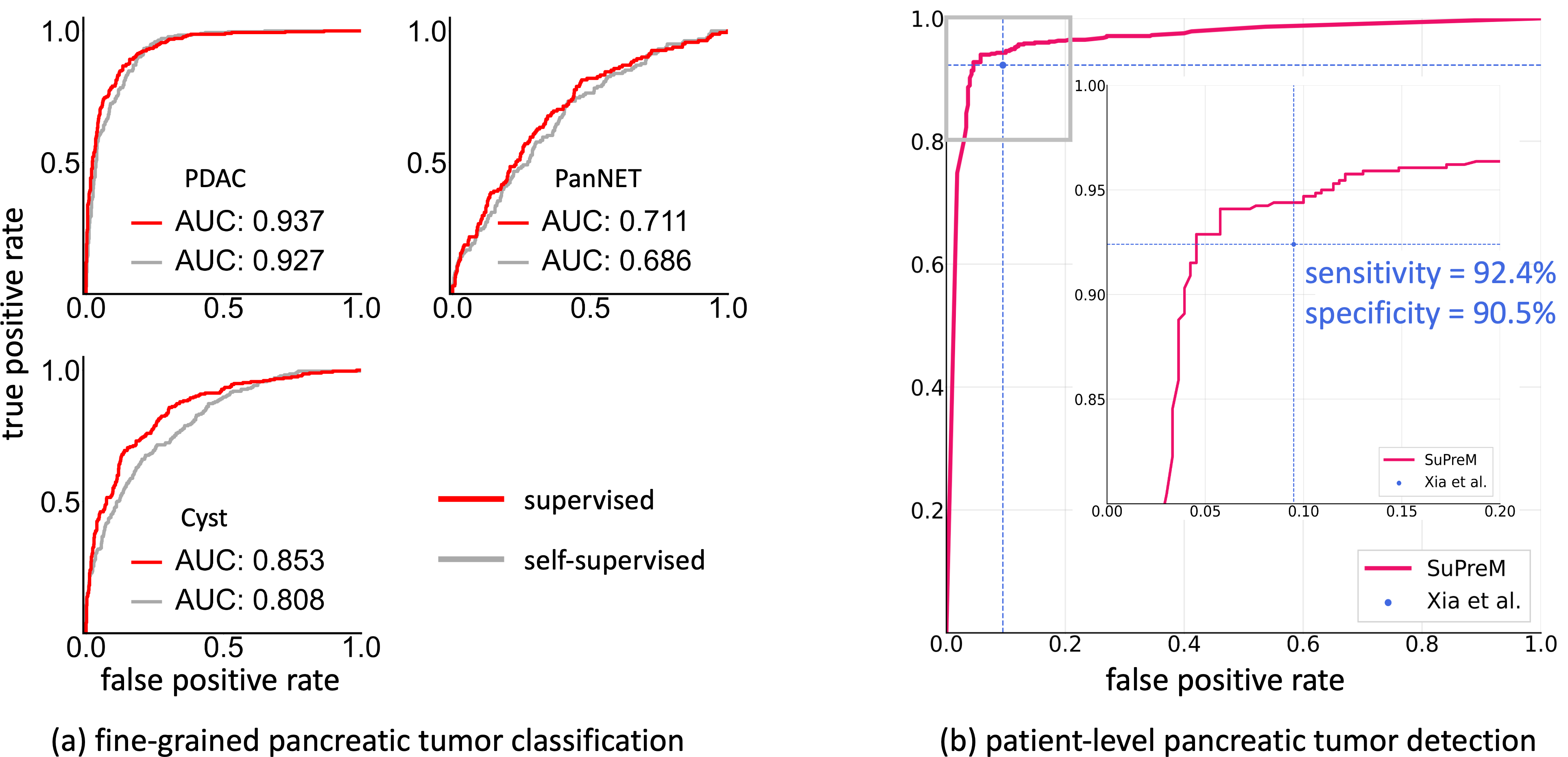}
    \caption{
    \textit{(a) Fine-tuning \oursupervisedmodel\ on fine-grained tumor classification involved the use of Receiver Operating Characteristic (ROC) curves to assess its transfer learning abilities in cross-task.} The task of identifying Cysts and PanNETs in the subset of \ourtestprivateAA\ presents unique difficulties for AI, as these lesions display a wider range of textural patterns compared to PDACs.  This variation in textural patterns is evident in the Area Under the Curve (AUC) values we recorded. Across all three pancreatic tumor sub-types, the supervised pre-training model (in \textcolor{red}{red}) demonstrates superior transfer learning ability than the self-supervised model \citep{tang2022self} (in \textcolor{gray}{gray}), showing its effectiveness in fine-grained tumor classification. \textit{(b) Fine-tuning \oursupervisedmodel\ on patient-level pancreatic tumor detection task.} The ROC curve of \oursupervisedmodel\ shows improved performance compared to the sensitivity of 92.4\% and specificity of 90.5\% as reported by \citet{xia2022felix}. We also provide a zoomed-in view of the ROC curve marked by a \textcolor{gray}{gray} box to visualize the superior tumor detection performance of \oursupervisedmodel.
    }
\label{fig:pancreatic_tumor_downstream_tasks}
\end{figure*}

\begin{figure}[t]
    \centering
    \includegraphics[width=1\linewidth]{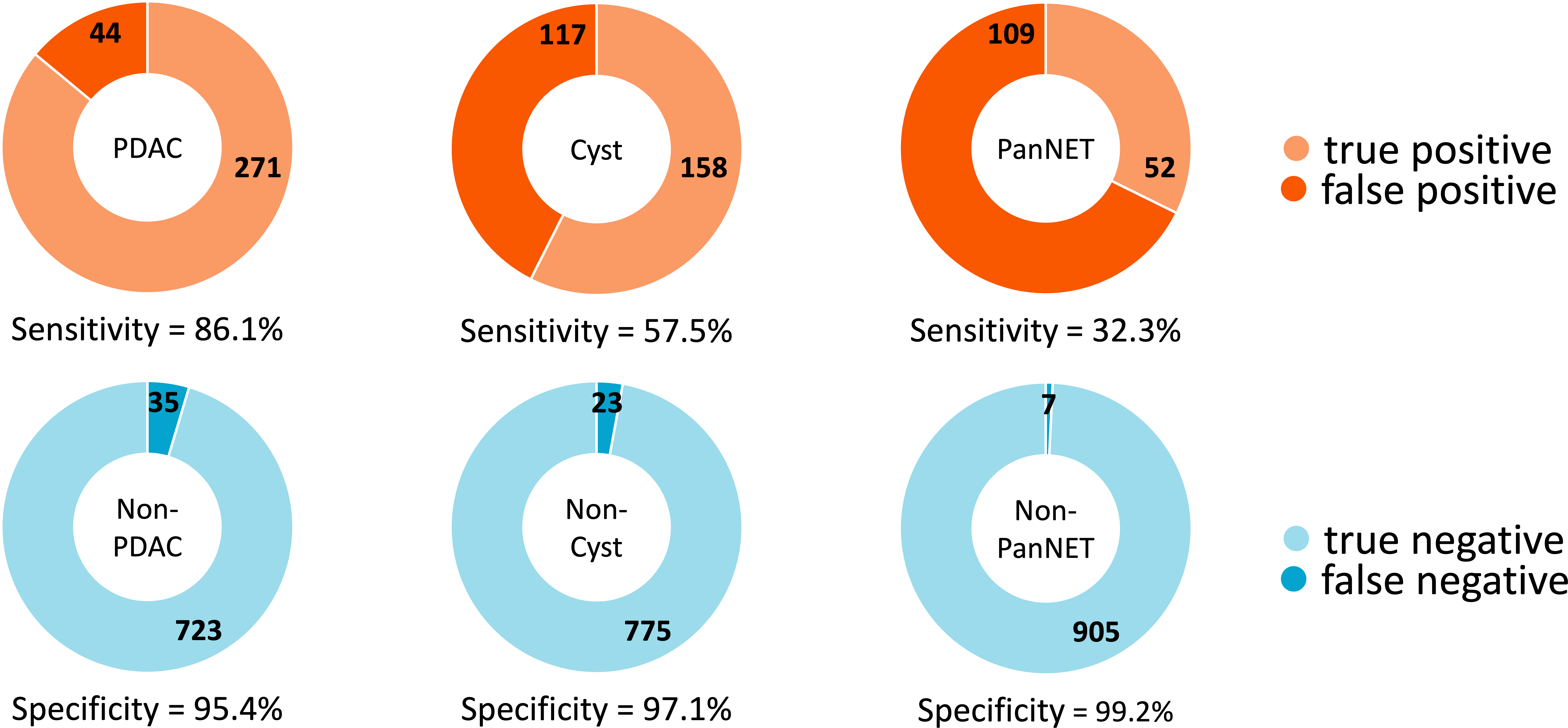}
    \caption{
    \textit{Fine-grained pancreatic tumor classification.} Notably, the transfer learning results demonstrate a sensitivity of 86.1\% and a specificity of 95.4\% in detecting PDAC. This performance exceeds the average radiologist's performance accuracy in identifying PDAC by 27.6\% in terms of sensitivity and 4.4\% in specificity, as indicated in \citet{cao2023large}.
    }
    \label{fig:tumor_identification}
\end{figure}

We have explored the transfer learning ability of \oursupervisedmodel\ for cross-task, where transferring from anatomical structures segmentation to fine-grained tumor identification. This shift represents a significant leap, as it involves moving from segmentation tasks to classification ones, which is inherently more challenging. The difficulty in evaluating fine-grained tumor classification largely stems from the limited annotations available in public datasets, often restricted to a few hundred tumors. Our \ourdataset\ addresses this challenge by using the subset of \ourtestprivateAA---containing 3,577 annotated pancreatic tumors, detailed in 1,704 PDACs, 945 Cysts, and 928 PanNets. As depicted in \figureautorefname~\ref{fig:pancreatic_tumor_downstream_tasks}(a), our findings indicate that supervised models, particularly \oursupervisedmodel, demonstrate superior transfer learning abilities to classification tasks compared to self-supervised models, as evidenced by their higher Area Under the Curve (AUC) in identifying each tumor subtype. The transfer learning results, as shown in \figureautorefname~\ref{fig:tumor_identification}, exhibit a sensitivity of 86.1\% and specificity of 95.4\% in detecting PDAC. These results exceed the average radiologist's accuracy in identifying PDAC, with a 27.6\% improvement in sensitivity and 4.4\% increase in specificity, according to the study \citep{cao2023large}. Additionally, \oursupervisedmodel\ also achieved better performance in patient-level tumor detection, surpassing the sensitivity of 92.4\% and specificity of 90.5\% reported in the study \citep{xia2022felix}, as shown by the ROC curve in \figureautorefname~\ref{fig:pancreatic_tumor_downstream_tasks}(b).

\section{Application II: Open Algorithmic Benchmarking}\label{sec:application_benchmark}

 We are organizing a medical segmentation challenge named \ourchallenge, using \ourdataset, hosted at the ISBI \& MICCAI 2024. \ourchallenge\ differs from many preexisting medical segmentation challenges because of the following features brought by our \ourdataset. \ourchallenge\ aims to encourage AI algorithms that are not just theoretically perform well, but also practically efficient and reliable in clinical settings.

\subsection{Features of \ourchallenge}

\subsubsection{A Large-scale Test Set}

To rigorously evaluate AI algorithms, we reserved a large-scale proprietary dataset from \ourtestprivateAA\ as the test set. This dataset comprises 1,000 CT volumes with high-quality annotations, used for external validation to ensure comprehensive and reliable evaluation while reducing overfitting risk. In future editions of \ourchallenge, we plan to expand this test set to 9,437 CT volumes and include 142 annotated anatomical structures, enhancing evaluation standards in medical imaging.

\subsubsection{A Pronounced Domain Shift}

\ourchallenge\ focuses on AI generalizability in real-world scenarios where a pronounced domain shift often occurs when applying AI algorithms trained on one hospital's data to a different hospital. \ourdataset\ is designed to address this problem. The test set (from \ourtestprivateAA\ private) features CT volumes with a high-resolution slice thickness of 0.5mm, significantly higher than the average 3mm resolution in the training set (from \ourtrainpublic\ public). This resolution difference presents an important challenge in generalizing AI algorithms from low-resolution training sets to high-resolution test sets. In \ourchallenge\ challenge, we reserve our test set and restrict how often participants can make their test submissions to prevent overfitting. Therefore, we recommend participants using external validation sets---such as \ourtestpublic\ (sourced from CH) which differs from the training set in patient demographics---to test the robustness of AI algorithms before making test submissions. Our comprehensive evaluation approach ensures that \ourchallenge\ serves as a rigorous and fair benchmark for medical segmentation tasks, similar to the role of COCO \citep{lin2014microsoft} in computer vision.

Samples that are independent and identically distributed with respect to the training data are dubbed IID or in-distribution. Examples are the CT volumes in test datasets constructed by randomly splitting a database into a training and an evaluation subset. Conversely, out-of-distribution (OOD) samples are not extracted from the training data distribution. Examples are CT volumes from hospitals not contributing to the training dataset. In medical imaging applications, AI accuracy on IID test samples may be much higher than performances on OOD data \citep{Geirhos2020Shortcut}. In such cases, AI may generalize poorly to real-world clinical scenarios (e.g., \citet{DeGrave2021AI}). We have access to test data drawn from hospitals that were never seen during training (e.g., JHH and YF in \tableautorefname~\ref{tab:abdomenatlas_makeup}). Therefore, in \ourchallenge, we evaluate generalizability across diverse clinical settings, which may encompass differences in patient demographics, equipment used, pathology prevalence and hospital protocols. Accordingly, \ourchallenge\ fosters the creation of fair, robust and reliable AI, which presents high performance well beyond its training domain. Preliminary results of the benchmark is presented in \tableautorefname~\ref{tab:preliminary_challenge}.

\begin{table}[h]
    \centering
    \scriptsize
    \caption{
    Preliminary results of testing AI algorithms on TotalSegmentator and JHH. We compare the out-of-distribution (OOD) performance of AI with the in-distribution (IID) performance of AI. The around 8\% difference of average DSC between OOD performance and IID performance and the even larger difference for hard-to-segment structures suggests the pronounced domain shift between \ourtrainpublic\ and other datasets.
    }\vspace{2px}
    \begin{tabular}{p{0.13\linewidth}P{0.15\linewidth}P{0.19\linewidth}|P{0.15\linewidth}P{0.15 \linewidth}}
    \toprule
    \multirow{2}{*}{class} & \multicolumn{2}{c|}{TotalSegmentator} & \multicolumn{2}{c}{JHH} \\
     & inf. \oursupervisedmodel & \citeauthor{wasserthal2022totalsegmentator} & inf. \oursupervisedmodel & \citeauthor{wang2019abdominal} \\
    \midrule
    spleen & 95.2{\tiny$\pm$0.0} & 98.4 & 95.0{\tiny$\pm$0.0} & 97.1 \\
    kidney~(R) & 92.5{\tiny$\pm$0.2} & 94.7 & 92.2{\tiny$\pm$0.0} & 98.4\\
    kidney~(L) & 89.0{\tiny$\pm$0.3} & 94.4 & 91.6{\tiny$\pm$0.1} & 96.8 \\
    gall bladder & 82.8{\tiny$\pm$0.2} & 84.5 & 83.6{\tiny$\pm$0.2} & 90.5\\
    liver & 94.7{\tiny$\pm$0.2} & 96.3 & 95.0{\tiny$\pm$0.3} & 98.0\\
    stomach & 85.2{\tiny$\pm$0.3} & 95.5 & 92.2{\tiny$\pm$0.1} & 95.2\\
    aorta & 75.6{\tiny$\pm$0.2} & 98.2 & 73.9{\tiny$\pm$0.3} & 91.8\\
    IVC & 74.2{\tiny$\pm$0.2} & 93.4 & 77.7{\tiny$\pm$0.4} & 87.0\\
    pancreas & 83.5{\tiny$\pm$0.2} & 89.4 & 79.0{\tiny$\pm$0.3} & 87.8\\
    \hline
    \textbf{average} & 85.9{\tiny$\pm$0.2} & 93.9 & 86.7{\tiny$\pm$0.2} & 93.6 \\
    \bottomrule
    \end{tabular}
    \label{tab:preliminary_challenge}
\end{table}

\subsubsection{Hard-to-segment Structures}

Although advanced AI algorithms have achieved impressive performance in some organ segmentation by reaching a Dice Similarity Coefficient (DSC) score around 0.98 (e.g., liver), they still face challenges in segmenting certain anatomical structures (e.g., small objects, structures with blurry boundaries, and tubular structures like the aorta). The deficiency of AI segmenting these structures was hard to reveal using preexisting datasets, due to either the absence of such classes or the presence of poor-quality annotations. Comparing with all preexisting datasets, our test set provides more comprehensive and high-quality annotations of these hard-to-segment structures. This allows us to more accurately test AI algorithms, helping to identify those that perform well on hard-to-segment structures and thereby advancing the field of medical image analysis.
\begin{figure*}[t]
    \centering
        \includegraphics[width=\linewidth]{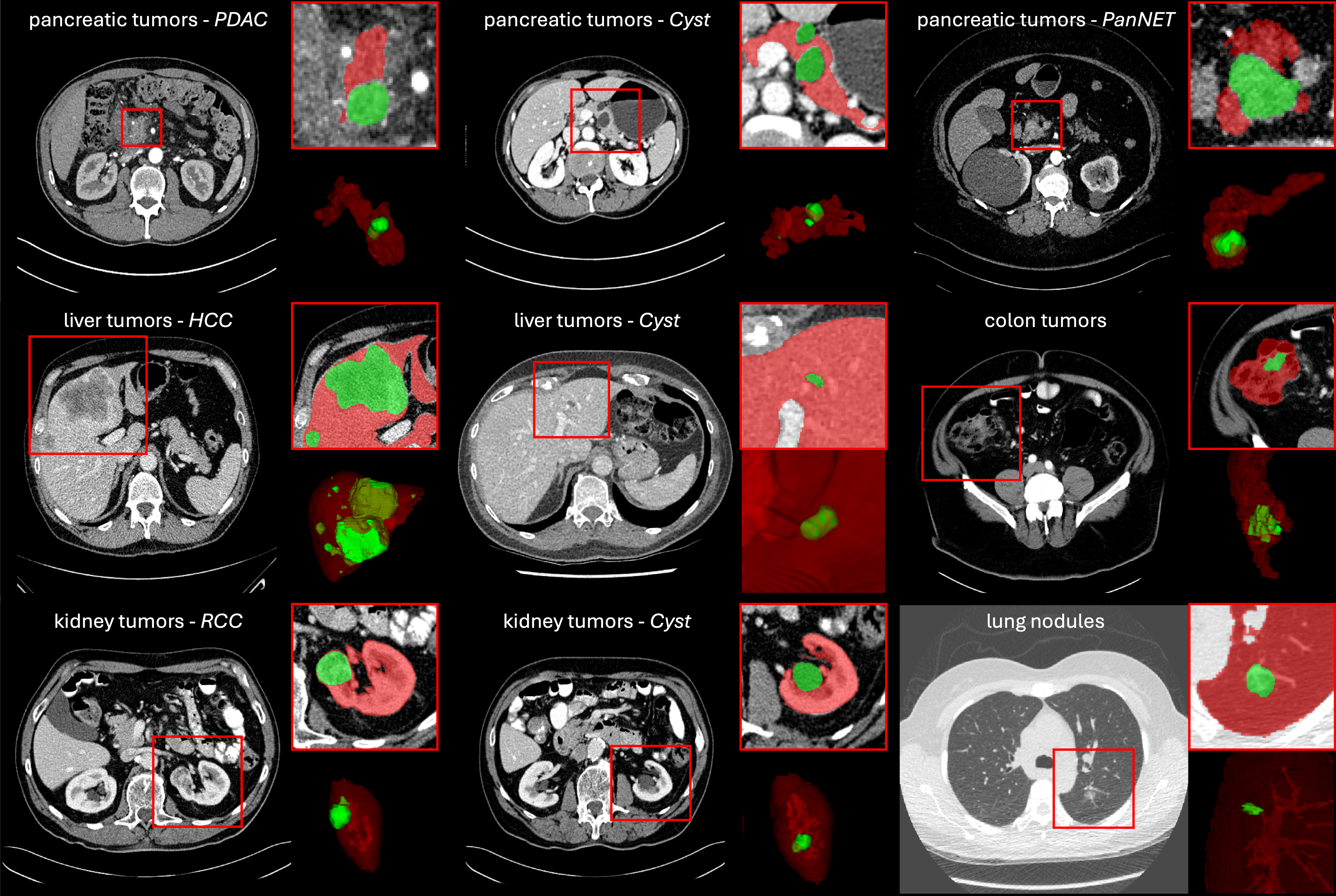}
        \caption{\textit{Tumor masks in \ourdataset.} We provide detailed per-voxel annotations of nine pathological structures in a subset of \ourdataset. This includes three sub-types of pancreatic tumors: pancreatic ductal adenocarcinoma (PDAC), pancreatic neuroendocrine tumors (PanNET), and pancreatic cysts; two sub-types of liver tumors: hepatocellular carcinoma (HCC) and cysts; two sub-types of kidney tumors: renal cell carcinoma (RCC) and cysts; as well as annotations for colon tumors and lung nodules.
        }
    \label{fig:tumor_annotation_examples}
\end{figure*}
\subsubsection{Inference Speed}

We emphasize the necessity for AI algorithms to efficiently process data while maintaining high performance. Therefore,
\ourchallenge\ introduces performance metrics and ranking paradigms that consider both segmentation accuracy and inference speed, which is especially crucial in clinical environments where timely decision-making can impact patient outcomes. In addition to speed,
the novel performance metrics in \ourchallenge\ also focus on the ability of algorithms to handle complex segmentation tasks, such as those involving pronounced domain shifts and hard-to-segment structures. These complex tasks often increase computational complexity, potentially slowing down inference times. Balancing this trade-off between speed and performance is an integral challenge in \ourchallenge, facilitating the development of AI solutions that are not only accurate at handling diverse scenarios but also efficient in their operation.

\section{Discussion \& Future Promises}\label{sec:discussion}

\subsection{Will Tumors in \ourdataset\ be Annotated?} 

While the semi-automatic annotation procedure in \ourdataset\ enables rapid scaling of annotations for various anatomical structures, it does not extend to tumor annotations. Annotating tumors is significantly more challenging due to their blurry boundaries, subtle intensity, small size, and varied conditions \citep{li2023early,hu2023label,hu2022synthetic,hu2023synthetic}. Additionally, there is no large-scale tumor dataset with detailed per-voxel annotations, making it difficult to develop strong AI algorithms for producing reasonable tumor pseudo annotations (Step 1 in our semi-automatic procedure in \S\ref{sec:step_by_step_procedure}). Poor pseudo annotations would significantly increase the revision workload for radiologists. Currently, we have made some progress in annotating tumors in \ourdataset. Nine types of annotated tumor examples are shown in \figureautorefname~\ref{fig:tumor_annotation_examples}. We have also generated 51.8K pseudo annotations for six tumor types predicted by a combination of state-of-the-art AI algorithms \citep{chen2024towards,lai2024pixel}. These pseudo annotations are preliminary and are pending validation by our collaborated radiologists and further verification through biopsy examinations. Based on our initial estimation, \ourdataset\ consists of 60\% normal (tumor-free) CT volumes and 40\% abnormal (tumor) CT volumes, which will be invaluable resources for cancer imaging in the future. 

To enhance \ourdataset\ and address its current limitations, we are initiating a comprehensive plan focusing on the integration of tumor annotations in three possible directions. Firstly, we plan to recruit more experienced radiologists to review and revise the tumor annotations. Secondly, the integration of pathology reports as weak annotations \citep{xiang2023squid,xiang2024exploiting,siddiquee2019learning}, derived from biopsy results, will be implemented to complement the radiologists' revisions, providing a multi-faceted approach to tumor annotations that seeks to minimize annotating biases and errors. Thirdly, we intend to utilize synthetic tumor data to generate a large collection of tumor examples and their corresponding precise segmentation masks for more effective AI training and validation. These three directions are aimed at significantly enriching \ourdataset\ with high-quality tumor annotations, thereby increasing its value for medical imaging AI development. In summary, creating annotated large-scale tumor datasets still requires significant collaborative efforts and innovative strategies from the medical imaging community.

\subsection{Segment Anything vs. Our Semi-automatic Approach}

An emerging research field consists of the creation of AI algorithms designed to segment arbitrary structures in medical images, according to prompts (e.g., bounding boxes) provided by the user \citep{ma2023segment}. Such algorithms were dubbed Segment Anything Models (SAM), after their counterparts in the field of natural image segmentation \citep{kirillov2023segment}. SAM can ``segment anything'' but does not know what is segmented. Therefore, the original SAM and its variants were limited when applied to medical images (especially for 3D volumetric data) \citep{ma2023segment,huang2024segment,guo2024comprehensive}. At the time we write this paper, SAM-based algorithms have not been integrated into standard medical annotation software (e.g., MONAI-LABEL and 3D Slicer). Once integrated, we expect SAM-based algorithms to improve the semi-automatic annotation procedure we described in this study: when revising annotations (Step 5, \S\ref{sec:semi_automatic_annotation}), radiologists can provide bounding boxes---or other weak annotations \citep{chou2024acquiring}---to SAM algorithms, and leverage the assistance of this interactive AI to revise an annotation more quickly \citep{zhang2024leveraging}. Moreover, 3D promptless SAM-based algorithms \citep{chen2023ma} represent a young but promising research field. If, in the future, these algorithms start to consistently surpass standard medical segmentation AI and become the new state-of-the-art for segmenting large CT scan datasets, they could be integrated with the nnU-Net, U-Net, and Swin UNETR in the semi-automatic annotation procedure we presented in \S\ref{sec:semi_automatic_annotation}.

\subsection{Are Labels Biased to Specific AI Architectures?}\label{sec:label_bias}

The \ourdataset\ annotations are created by a synergy between humans and AI. It is possible for annotations produced by a specific AI architecture to present patterns that are characteristic of the architecture. Thus, a U-Net may be able to better fit annotations that were originally created by another U-Net. However, three diverse architectures contribute to creating the \ourdataset\ annotations: Swin UNETR, U-Net, and nnU-Net. This procedure increases annotation diversity and accuracy (errors characteristic of one architecture are effectively diluted by the others), and it prevents any single architecture from dominating
the annotation process (by averaging the predictions from these three architectures), decreasing potential annotation bias towards a specific architecture. Additionally, a subset of \ourtestprivateAA\ was manually annotated. This characteristic further reduces the influence of potential architecture-specific bias on AI test performances reported for \ourdataset.

\subsection{Future Promises}\label{sec:future_promise}

We have made 9,262 annotated CT volumes from 88 hospitals publicly available for training. Moreover, a significant part of our test set is also public(\ourtestpublic, \tableautorefname~\ref{tab:abdomenatlas_makeup}). Our private test set, \ourtestprivateAA, encompasses 15 hospitals and 9,437 CT volumes. Of these 9,437 CT volumes, 8,189 (86.7\%) are reserved for fair evaluation which keeps both CT volumes and annotations private (JHH, CirrosisPro and YF, \tableautorefname~\ref{tab:abdomenatlas_makeup}).
The remaining 1,248 CT volumes in \ourtestprivateAA\ are public, but the annotations we created for them are currently private. These 1,248 volumes are essential for boosting the data diversity of \ourtestprivateAA, increasing its number of hospitals from 9 to 15. A large, diverse, and private dataset, like \ourtestprivateAA, is an important asset for performing third-party evaluation in algorithmic benchmarks. First, the unprecedented scale of \ourtestprivateAA\ increases the statistical significance of our \ourchallenge\ challenge results. Second, AI performance may strongly vary across diverse hospitals \citep{svanera2024fighting,lin2024shortcut,huang2023eval}. Accordingly, with 15 hospitals in \ourtestprivateAA, we can evaluate how well AI algorithms generalize to multiple clinical settings, and compare performances on hospitals that were seen during training, to hospitals that were never seen. Third, the private and inaccessible nature of the \ourtestprivateAA\ test set is essential to ensure that the teams participating in our challenges cannot use this test data for training, nor overfit the test set, thus guaranteeing the integrity of the \ourchallenge\ benchmarks' results. We will organize multiple challenges using \ourdataset. After a series of challenges, we plan to publish part of \ourtestprivateAA\ progressively.

To address the current limitations of \oursupervisedmodel, we plan to leverage its strengths--the extensive training on 25 anatomical structures across numerous CT volumes. This existing proficiency creates a strong foundation for enhancing \oursupervisedmodel's capabilities in identifying pathological structures. Because a well-trained organ-specific foundation model can provide better accuracy and interoperability as well as significantly reduce the amount of labeled data required for new tasks, particularly tumor-related tasks, as it has already learned relevant organ features and location prior for tumors from the previous training \citep{zhang2023challenges}.
Furthermore, we are dedicated to advancing our fine-tuning processes and integrating more advanced AI architectures and adapters to extend \oursupervisedmodel's proficiency in tumor identification and adaptability to multimodal medical imaging. This strategic enhancement is designed to optimize the performance of \oursupervisedmodel~in addressing complex medical challenges, thereby fulfilling its promise as a transformative foundation model in medical diagnosis and treatment.

Lastly, the scope of \ourdataset\ and AI algorithms trained on \ourdataset\ are currently limited to a single imaging modality---CT. The medical imaging field recognizes a substantial difference in the representation of anatomical and pathological structures across different modalities, such as CT and magnetic resonance imaging (MRI). This variation presents a challenge in developing a universal dataset that is modality-comprehensive. It underscores a potential area for future expansion and highlights the necessity to extend \ourdataset\ to a broader range of imaging modalities, thereby enhancing the robustness and applicability of AI algorithms developed using \ourdataset.

\section*{Acknowledgments}

This work was supported by the Lustgarten Foundation for Pancreatic Cancer Research and the Patrick J. McGovern Foundation Award. P.R.A.S.B. thanks the funding from the Center for Biomolecular Nanotechnologies, Istituto Italiano di Tecnologia (73010, Arnesano, LE, Italy). We thank The FELIX Team at Johns Hopkins Medicine \citep{park2020annotated} for collecting and annotating the JHH dataset; thank Hualin Qiao for reviewing and revising the annotation in \ourtrainpublic; thank Yu-Cheng Chou, Angtian Wang, Yaoyao Liu, Yucheng Tang, and Qi Chen for organizing the \ourchallenge\ competition at ISBI \& MICCAI-2024; thank Junfei Xiao, Jieneng Chen for their constructive suggestions at several stages of the project; thank Jaimie Patterson for disseminating our research findings. The content of this paper is covered by patents pending.

\bibliographystyle{model2-names.bst}\biboptions{authoryear}
\bibliography{refs,zzhou}

\end{document}